\documentclass[prx,aps,showpacs,twocolumn,preprintnumbers,
amsmath,amssymb,superscriptaddress,longbibliography]{revtex4-2} 
\usepackage[english]{babel}
\usepackage{amsmath}
\usepackage{amssymb}
\usepackage{amsfonts}
\usepackage{physics}
\usepackage{graphicx}
\usepackage[colorlinks=True,linkcolor=red,citecolor=Red,urlcolor=Red]{hyperref}
\usepackage{bookmark}
\usepackage[dvipsnames]{xcolor} 
\usepackage{braket}
\usepackage[normalem]{ulem}
\usepackage[normalem]{ulem}
\usepackage{nicefrac}
\usepackage{bm}
\usepackage{bbm}
\usepackage{braket}
\usepackage{slashed}
\usepackage{mathdots}
\usepackage[normalem]{ulem}
\usepackage{array}
\usepackage{makecell}
\usepackage{xspace}

\usepackage{cmap}
\usepackage{setspace}
\usepackage{amstext}			
\usepackage{amsfonts}
\usepackage{bbold}
\usepackage{graphicx}
\usepackage{subfigure}
\usepackage{wrapfig}
\usepackage{esint}
\usepackage{hyperref}
\usepackage{multirow}
\usepackage{printlen}
\usepackage{placeins}
\usepackage{comment}
\usepackage{bm}
\usepackage{tikz-cd}
\usepackage{tkz-berge}
\usepackage[boxsize=1em]{ytableau}
\usepackage[english]{babel}
\usepackage{fancyhdr}
\usepackage[left=1in, right = 1in, top = 1in, bottom = 1in]{geometry}

  \DeclareUnicodeCharacter{2212}{-}

\allowdisplaybreaks

\newcommand{\TopTimeNet}{Top{\textsc{Time}}Net\xspace}

\begin{document}

 \title{
\TopTimeNet: Topologically-assisted time-series classification model
 }
  \author{Sharareh Sayyad}
 \affiliation{Department of Mathematics and Statistics, Washington State University, \\Pullman, Washington 99164-3113, USA}
 \email{sharareh.sayyad@wsu.edu}
 \author{Sophia Bazzi}
\affiliation{European Molecular Biology Laboratory, EMBL Hamburg, c/o DESY, Building 25A, Notkestraße 85, 22603 Hamburg, Germany}
\email{sophia.bazzi@embl-hamburg.de}

\begin{abstract}
Distinguishing periodic from chaotic dynamics in a time series is a fundamental challenge in both physics and engineering. 
Yet, end-to-end learned architectures must discover both a representation and a decision boundary from data, at substantial cost. We introduce \TopTimeNet{}, which decouples these tasks: a fixed, non-learned stage extracts a $42$-dimensional geometric and topological descriptor from Takens delay embeddings and persistent homology, and a lightweight learnable stage performs classification. On a benchmark of $49$ nonlinear dynamical systems, a $1{,}638$-parameter configuration matches the mean accuracy of one with $33\times$ more trainable parameters. 
Additionally, this approach delivers mean accuracy comparable to convolutional neural networks and surpasses the average performance of converged Transformer models, while requiring three to four orders of magnitude fewer trainable parameters.
Robustness also depends sharply on where noise is introduced: \TopTimeNet{} degrades gracefully under perturbations to its precomputed features, but degrades sharply when noise
is introduced into the raw signal and the full feature-extraction pipeline is recomputed, showing that robustness to perturbations of the precomputed features does not imply robustness of the complete raw-signal-to-prediction pipeline.
These results show that decoupling fixed geometric and topological feature construction from a lightweight discriminative stage can achieve comparable classification accuracy with substantially fewer trainable parameters.
\end{abstract}

\maketitle

\section{Introduction}\label{sec:introduction}

Nonlinear dynamical systems can exhibit qualitatively different regimes of motion, ranging from periodic and quasiperiodic oscillations to chaos. In many experiments, however, the governing equations are unknown, or only a subset of the state variables is accessible. The dynamical regime must then be inferred from a finite time series. This inference problem is nontrivial: chaotic and stochastic processes can both generate irregular signals, and finite data length or measurement noise may obscure the distinction between them~\cite{Provenzale1992, Tsonis1992, Cencini2000}.

Classical nonlinear time-series analysis offers several tools to address this problem. The largest Lyapunov exponent quantifies sensitivity to initial conditions, while correlation dimension estimates the fractal scaling of the attractor, giving a measure of its effective dimensionality~\cite{Grassberger1983b}. Power spectra and autocorrelation functions provide information about temporal structure. 
These methods are valuable, but each probes a particular property of the signal. Correlation-dimension estimates, for example, are well known to be sensitive to observational noise~\cite{casaleggio1995,argyris1998}.
More generally, finite data and limitations in phase-space reconstruction can complicate the interpretation of nonlinear time-series measures~\cite{Cencini2000}.
For this reason, a reliable analysis often requires several diagnostics rather than a single scalar measure.

Machine-learning (ML) methods offer a complementary strategy. Rather than choosing a specific dynamical indicator in advance, they learn features useful for classification directly from data. Deep neural networks have been used to distinguish chaotic from non-chaotic time series across discrete and continuous dynamical systems~\cite{boulle2020classification}.
Researchers have also studied data-driven learning methods to distinguish deterministic chaos from stochastic behavior~\cite{Zanin2022,choi2026learning}.
Many ML approaches operate directly on the raw time series~\cite{Wang2017, IsmailFawaz2019}. Convolutional neural networks (CNNs) can learn local and multiscale temporal patterns through convolutional filters~\cite{IsmailFawaz2020}. Transformer-based models employ attention to capture dependencies and interactions across longer time scales~\cite{Wen2023}. 
While these architectures are flexible, they often require many trainable parameters and substantial computational resources to learn a useful representation from scratch.
For dynamical systems, we can instead extract relevant geometric and topological structure before learning begins, allowing the learnable stage to focus on projecting, combining, and classifying these features.
This raises a central question we address in this work: how much learnable capacity is needed once we make such structure explicit?

Topological data analysis (TDA) offers a way to construct such features. A scalar time series can first be mapped to a point cloud in a reconstructed phase space using delay-coordinate embedding~\cite{takens2006}. The resulting point cloud retains geometric information about the observed trajectory. Persistent homology then tracks topological features across different length scales. Connected components are described by zeroth homology, while loops are described by first homology. Persistence-based summaries have been used to distinguish dynamical states such as periodic and chaotic behavior~\cite{Myers2019}. 
Tempelman and Khasawneh further report that persistence-based chaos detection remains tolerant to moderate observational noise in several benchmark systems~\cite{Tempelman2020}. In Sec.~\ref{sec:robustness}, we separately examine robustness when noise is introduced before and after feature extraction.

Several representations have been developed to adapt persistent homology information for statistical learning. Persistence landscapes provide functional summaries of persistence diagrams~\cite{bubenik2015statistical}. Persistence images map diagrams to stable finite-dimensional vectors~\cite{adams2017persistence}. Signature-based feature maps provide another representation of persistence barcodes for statistical learning~\cite{chevyrev2018persistence}. Persistent entropy has also been used to construct stable summary functions that incorporate information from Betti curves~\cite{atienza2020stability}. For time-series data, Umeda combined an engineered topological feature with a CNN-inspired learning architecture~\cite{umeda2017time}. Karan and Kaygun later developed persistent-homology-based pipelines for univariate time-series classification~\cite{Karan2021}.

The use of topological features in time-series learning is thus already well established, and prior work has combined several persistence-derived summaries within a single pipeline~\cite{Karan2021}. 
For instance, Karan and Kaygun summarize persistence diagrams using diagram distances, persistent entropy, and scalar norms of Betti curves and persistence landscapes; their pipeline does not include persistence images or geometric statistics computed directly from the embedding point cloud.

Our study focuses on how these complementary geometric and topological feature groups can be organized within a compact model. We introduce \TopTimeNet{}, an architecture for classifying periodic and chaotic time series. Each segmented time series is first mapped to a delay-coordinate point cloud. We compute geometric features directly from this point cloud and obtain additional topological descriptors from persistent homology. The resulting $42$ features are divided into five groups: geometric, entropy, lifetime, Betti-curve, and persistence-image features. This feature-extraction stage contains no learned weights, although it depends on fixed analysis settings such as the embedding parameters and persistence-image resolution.

The five feature groups are then passed to the learnable part of \TopTimeNet{}. Each group is projected into a common embedding space. A fusion layer combines the projected features, and a classification head produces the final prediction. Learning therefore focuses on interactions between the extracted feature groups rather than reconstructing the full representation from the raw signal.

This design enables us to investigate the amount of trainable capacity needed for the classification task. We compare compact and higher-capacity \TopTimeNet{} configurations selected through the same hyperparameter search. The two configurations differ by a factor of approximately $33$ in parameter count, yet achieve nearly identical mean classification accuracy. We also compare \TopTimeNet{} with CNN and Transformer baselines trained directly on the raw time series.

We additionally examine the response of \TopTimeNet{} to Gaussian perturbations introduced at two distinct points in the pipeline: directly on the precomputed feature vectors, and on the raw time series itself, with the full delay-embedding and persistent-homology computation repeated on the corrupted signal.

Under the tested perturbations, accuracy degrades more sharply when noise is introduced in the raw time series and the features are recomputed than when noise is applied directly to the precomputed feature vectors. This shows that robustness to feature-level perturbations does not imply robustness of the complete raw-signal-to-prediction pipeline. 

The remainder of this paper is organized as follows. Section~\ref{sec:method} describes the dataset, the geometric and topological feature-extraction pipeline, the \TopTimeNet{} architecture, and the training procedure. Section~\ref{sec:example} illustrates the extracted features using the double pendulum as an example. Section~\ref{sec:results} presents the classification results, the response to feature-level and raw-signal noise, and comparisons with CNN and Transformer baselines. Finally, Section~\ref{sec:conclusion} summarizes the main findings and outlines directions for future work.

\section{Methodology}\label{sec:method}

In this section, we present details on dataset generation, the \TopTimeNet{} architecture, and the procedures used to train and evaluate the models for time-series classification.

\subsection{Dataset generation}\label{sec:data_generation}

To construct the time-series dataset, we used a catalog of 49 nonlinear dynamical systems from the Teaspoon library~\cite{Khasawneh2025}. This catalog includes various discrete maps, such as the logistic map; dissipative flows, including the Lorenz attractor~\cite{lorenz2017} and the driven pendulum; and conservative flows, such as the H\'enon-Heiles system~\cite{henon1964}.

We simulated each system in the catalog in both periodic and chaotic parameter regimes, where available, yielding $96$ (dataset, state) pairs, up to a fixed number of time steps, $N$, usually set to $100{,}000$. For systems described by $n$ coupled differential equations, the simulations produce $n$-dimensional trajectories. We treat each component of these solutions separately to ensure that each time series in our dataset is univariate, yielding $239$ individual signals across the $96$ pairs. To further increase the number of samples for each dynamical regime, we segment each of these signals into segments of length $T$. With $N=100{,}000$ and $T=1{,}000$, each signal produces $100$ segments per signal, leading to a total of $23{,}900$. However, $600$ of these segments are discarded for being identically zero. The remaining $23,300$ segments are unbalanced between classes, with $12,000$ labeled as chaotic and $11,300$ as periodic. To achieve balance, we undersample the majority class (chaotic) to match the minority class, resulting in a final, balanced dataset of $22,600$ time series, equally divided between the periodic and chaotic classes.

\subsection{\TopTimeNet model}\label{sec:alg}

\begin{figure*}
    \centering
    \includegraphics[width=0.95\linewidth]{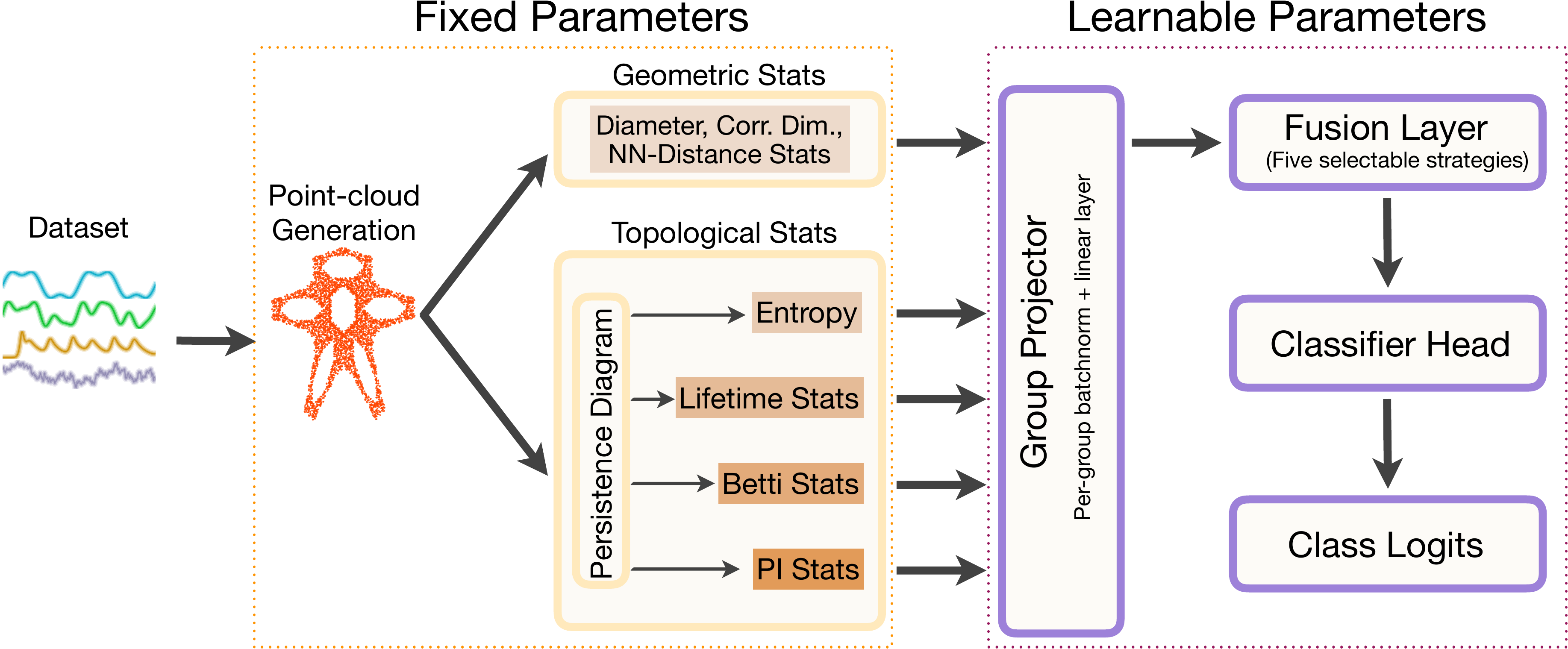}
    \caption{\textbf{Schematic illustration of the \TopTimeNet algorithm.} The segmented time-series data first pass through a non-trainable feature-extraction stage of the algorithm, where point clouds are generated and geometric and topological statistics are extracted. This process creates a complete feature set that will be used in subsequent steps. The five groups of extracted features are then processed through a Group Projector and combined using a Fusion Layer. Finally, the combined features are classified using a Classifier Head to produce the final class logits.}
    \label{fig:alg}
\end{figure*}

Our \TopTimeNet model, schematically illustrated in Fig.~\ref{fig:alg}, consists of two main components: feature extraction and learnable fusion and classification. We present each component below.

\subsubsection{Feature extraction}\label{sec:alg-features}

Starting from a segmented time series $x(t) \in \mathbb{R}^T$, we employ Takens' delay embedding~\cite{takens2006} to map the signal into a reconstructed phase space. The corresponding delay-coordinate vectors are
$x_{\rm embd}(t) = [x(t), x(t+\tau), \ldots, x(t+(d_{\rm emb}-1)\tau)]$, and their collection forms the point cloud.
Here, $d_{\rm emb}$ is the embedding dimension and $\tau$ is the time delay. As a result of this process, $N_{\rm pt}=T-(d_{\rm emb}-1)\tau$ points, each with dimension $d_{\rm emb}$, are created in each point cloud.
For the results reported here, we set $d_{\rm emb}=2$ and $\tau=20$.
These embedding parameters are fixed throughout the hyperparameter search and final evaluation and are not included in the random search described in Sec.~\ref{sec:hyp-search}.

Once the point clouds for all segmented time series have been constructed, the pipeline splits into two parallel branches that extract geometric and topological features.

The geometric branch operates directly on the point cloud. Here, pairwise Euclidean distances between all $N_{\rm pt}$ points in the point cloud are computed using

\begin{align}
    D_{ij}= \| x_i - x_j \|_2
    =\sqrt{\sum_{k=1}^{d_{\rm emb}}(x_{i,k}-x_{j,k})^2},
\end{align}
where $i,j \in \{1, \ldots , N_{\rm pt}\}$.

Given the pairwise distance matrix $D$, we extract four geometric features. The first is the diameter of the point cloud,
\begin{align}
d_{\rm pt} = \max_{1\leq i,j \leq N_{\rm pt}} D_{ij}.
\end{align}

We then compute the nearest-neighbor distance of every point in the point cloud as
\begin{align}
    D_{nn,i} = \min_{j \neq i} D_{ij},
\end{align}
for $i \in \{1, \ldots , N_{\rm pt}\}$. The mean and standard deviation of the nearest-neighbor distances $D_{nn} \in \mathbb{R}^{N_{\rm pt}}$ constitute the next two geometric features.

Finally, we construct a Grassberger--Procaccia correlation-dimension proxy~\cite{grassberger1983,Theiler1987}. The correlation integral at distance $r$ reads
\begin{equation}
    C(r) =\frac{2}{N_{\rm pt}(N_{\rm pt}-1)} \sum_{i < j} \mathbb{1}(D_{ij}<r),
\end{equation}
where $\mathbb{1}$ denotes the indicator function. For a fractal set, $C(r)$ is expected to exhibit power-law scaling over an appropriate scaling regime as $C(r) \propto r^{d_{\rm cr}}$, where $d_{\rm cr}$ denotes the correlation dimension. Given two radii $r_s$ and $r_l$, one may approximate $d_{\rm cr}$ as
\begin{align}
    d_{\rm cr} \approx
    \frac{\log(C(r_l)/C(r_s))}
         {\log(r_l/r_s)}.
\end{align}

We choose $r_s$ and $r_l$ to be the 10th and 50th percentiles of the pairwise-distance distribution, respectively, providing a reproducible, data-adaptive pair of radii. Because the slope is evaluated between these two radii rather than fitted over an identified scaling regime, we treat $d_{\rm cr}$ as a correlation-dimension proxy rather than a full correlation-dimension estimate. This proxy is the final feature extracted by the geometric branch.

The second branch computes Vietoris--Rips persistent homology of each point cloud for a fixed number $H_{\rm max}$ of homology dimensions, yielding one persistence diagram for each $H_k$, with $k=0,\ldots,H_{\rm max}-1$~\footnote{For computational reasons, we set $H_{\rm max}=2$ and therefore retain $H_0$ and $H_1$.}.

The Vietoris--Rips construction builds a nested sequence of simplicial complexes by growing a ball of radius $r$ around every point in the cloud and increasing $r$ from zero. Whenever two balls overlap, an edge is added between their centers, and whenever all the edges among a set of points are present, forming a triangle, tetrahedron, or higher-dimensional simplex, that simplex is added as well, regardless of whether the corresponding balls share a common intersection.
At $r=0$, the complex consists only of isolated points, and as $r$ increases, edges and faces are added progressively, connecting the point cloud into an increasingly dense complex.

As $r$ grows, topological features across dimensions appear and disappear: connected components ($H_0$), loops ($H_1$), and, more generally, $k$-dimensional holes ($H_k$) for $k$ up to $H_{\rm max}-1$. Each such feature $i$ is born at the smallest radius $r_{\rm b}$ at which it first appears, and dies at the radius $r_{\rm d}$ at which it merges with an older feature or is filled in by higher-dimensional simplices. Because two balls of radius $r$ first overlap once their centers are separated by distance $2r$, we record these events in terms of the underlying pairwise-distance threshold rather than the radius itself, setting $b_i = 2r_{\rm b}$ and $d_i=2r_{\rm d}$. The persistence diagram for a given homology dimension is then the collection of all such pairs $(b_i,d_i)$ observed as $r$ is swept.

The single essential class in $H_0$ (the connected component that never dies) has $d_i=\infty$ in the standard construction; we cap its death time once, at the largest finite death value observed across all tracked homology dimensions for that point cloud, before computing any downstream statistics. This ensures every reported quantity, including the entropy, lifetime, Betti-curve, and persistence-image features described below, is finite-valued by construction~\footnote{Because the capped death value depends on the largest finite death observed for a given point cloud, the resulting lifetime assigned to the essential $H_0$ class is an implementation-dependent finite value and should not be interpreted as a topological invariant of the underlying attractor.}.

The lifetime (or persistence) of feature $i$ is the duration over which it survives, $l_i=d_i-b_i$; 
long-lived features represent structure that persists over a broader range of filtration scales, whereas short-lived features can arise from noise or small-scale structure.
Normalizing the lifetimes across all features in a diagram gives a probability distribution $p_i = l_i / \sum_j l_j$, which is used below to summarize the distribution of persistence across topological features through its entropy.

In \TopTimeNet{}, each persistence diagram at a fixed homology dimension $k$ is passed through four branches for further feature extraction. We describe these four branches below.

\paragraph{Entropy branch.}
The Shannon entropy of the lifetimes is computed at each homology dimension as
\begin{equation}
    H_{\rm ent} = -\sum_{i} p_i \log p_i,
\end{equation}
where $p_i = l_i / \sum_j l_j$ is the normalized lifetime associated with feature $i$. 
The entropy $H_{\rm ent}$ summarizes how the total persistence is distributed among the topological features~\cite{atienza2019persistent}. Low $H_{\rm ent}$ indicates that the persistence is concentrated in a small number of dominant features, as may occur for periodic signals. In contrast, high $H_{\rm ent}$ indicates that the persistence is distributed more evenly across multiple features, as may occur for chaotic signals.

\paragraph{Lifetime branch.} 
Let $l = (l_1, \ldots, l_L)$ denote the vector of lifetimes at a given homology dimension. We compute five summary statistics from $l$. The \emph{maximum lifetime}, $\max_i(l_i)$, indicates the presence of the most persistent topological feature in the simplicial complex, while the \emph{total lifetime}, $\sum_i l_i$, provides a measure of the overall persistence of the topological features.

The \emph{dominance ratio}, $\max_i(l_i) / \sum_j l_j$, estimates the fraction of the total persistence carried by the most persistent feature. A value close to $1$ indicates that one feature dominates, as may occur for clean periodic signals, while a value close to $0$ indicates that persistence is spread across many features, none of which dominates, as may occur for chaotic signals.

The \emph{coefficient of variation}, ${\rm std}(l)/{\rm mean}(l)$, measures the relative spread of lifetimes. Finally, the \emph{number of significant lifetimes} is defined as the count of features satisfying $l_i > 0.05 \times \max_j(l_j)$~\footnote{The factor $0.05$ defines a fixed relative threshold used throughout the analysis. Here, ``significant'' refers only to this thresholding criterion and does not imply statistical significance.}. This provides a proxy for the number of prominent topological features by excluding features with lifetimes that are small relative to the maximum lifetime.

\paragraph{Betti curve branch.}
Given the set of birth--death pairs $(b_i,d_i)$ from the persistence diagram at a fixed homology dimension $k$, we define the Betti curve, which counts the number of topological features alive at a given Vietoris--Rips filtration threshold $\varepsilon$, as
\begin{align}
    \beta(\varepsilon)
    =
    \sum_i \mathbb{1}(b_i \leq \varepsilon < d_i).
\end{align}
The function $\beta(\varepsilon)$ is piecewise constant, increasing by $1$ at each birth and decreasing by $1$ at each death, or by more than $1$ when several features share the same birth or death threshold.

To obtain a fixed-dimensional representation that is comparable across point clouds, we sample the Betti curve at a fixed number of filtration thresholds, $n_{\rm bin}=50$. Because the point clouds have different diameters, the thresholds are placed on a diameter-scaled grid,
\begin{align}
    \varepsilon_j
    =
    \frac{j}{n_{\rm bin}-1}d_{\rm pt},
    \qquad
    j=0,\ldots,n_{\rm bin}-1,
\end{align}
where $d_{\rm pt}$ is the diameter of the point cloud. Thus, the sampled filtration locations correspond to normalized positions $\varepsilon_j/d_{\rm pt}\in[0,1]$.

Given the sampled Betti curve
$\beta=(\beta_0,\ldots,\beta_{n_{\rm bin}-1})$,
we summarize it with six statistics. The \emph{maximum}, $\max_j\beta_j$, records the largest number of topological features alive simultaneously at any sampled filtration threshold. The \emph{mean}, $\overline{\beta}$, is the average value of the Betti curve across the sampled thresholds, while the \emph{standard deviation}, ${\rm std}(\beta)$, summarizes the variation in feature counts across the filtration. The \emph{peak position} is defined as
\begin{align}
    \frac{j^*}{n_{\rm bin}-1},
    \qquad
    j^*=\operatorname*{arg\,max}_{j}\beta_j,
\end{align}
and gives the normalized filtration location at which the sampled Betti curve reaches its maximum. Because Betti curves are piecewise constant, this maximum can be attained on a plateau spanning several consecutive $j$; our implementation follows the standard \texttt{argmax} convention of returning the first such index, so $j^*$ identifies the earliest filtration threshold at which the maximum is reached, rather than, e.g., the plateau's midpoint or last index.
The \emph{turning-point count} is obtained by first taking the discrete difference of the curve,
$\Delta\beta_j = \beta_{j+1} - \beta_j$ for $j = 0, \ldots, n_{\rm bin}-2$, and then counting the indices $j=0,\ldots,n_{\rm bin}-3$ for which two consecutive differences $\Delta\beta_j$ and $\Delta\beta_{j+1}$ have opposite signs, i.e.\ $\Delta\beta_j \cdot \Delta\beta_{j+1} < 0$.
These sign changes identify turning points of the Betti curve and provide a proxy for how oscillatory or multi-modal the curve is. Because Betti curves are piecewise constant, this adjacent-sign-change criterion does not detect turning points separated by one or more zero-valued differences (plateaus); it therefore provides a conservative lower bound on the number of true direction changes in the curve, rather than an exhaustive count.

Finally, the \emph{bimodality coefficient},
\begin{align}
    {\rm BC} = \frac{m^2+1}{\kappa},
\end{align}
provides a skewness--kurtosis-based summary of the distribution of the sampled $\beta$ values.
Here $m$ is the skewness (standardized third moment) and $\kappa$ is the standard, non-excess kurtosis (standardized fourth moment) of $\beta$.
Larger values of BC are sometimes interpreted as being compatible with bimodality, but BC is a heuristic rather than a direct test of the number of modes and can also be large for strongly skewed unimodal distributions.

\paragraph{Persistence image branch.}
An alternative representation of the set of birth--death pairs $(b_i,d_i)$ at a fixed homology dimension is obtained by treating them as a kernel-smoothed surface rather than a set of isolated points. 
Following the persistence-image construction~\cite{adams2017persistence},
each pair is first mapped to birth--persistence coordinates, $(b_i,\ell_i)$ with $\ell_i=d_i-b_i$, and we define a weighted density surface over the birth--persistence plane as
\begin{align}
    \rho(x,y)
    &= \sum_i
    \frac{w(\ell_i)}{2\pi\sigma^2}
    \exp\left(
    -\frac{(b_i-x)^2+(\ell_i-y)^2}{2\sigma^2}
    \right),\\
    w(\ell)
    &= \ell,
\end{align}
where $\sigma>0$ is the width of the Gaussian kernel (we set $\sigma=0.05$). The weight $w(\ell)$ increases linearly with lifetime, so that longer-lived topological features contribute more strongly to the surface than short-lived ones.

Each pixel of the resulting $n_{\rm pi}\times n_{\rm pi}$ grid 
(we set $n_{\rm pi}=15$)
is assigned the integral of $\rho$ over that pixel's area, rather than the value of $\rho$ evaluated at the pixel center, yielding the \emph{persistence image} (PI). Before it can be evaluated, the grid must be calibrated: its birth and persistence bounds are set once from a training set of diagrams at a given homology dimension and then held fixed for all subsequent evaluations, ensuring that no information from validation or test samples enters the bound calibration. This ensures that all persistence images at a given homology dimension are expressed on an identical pixel grid and are therefore directly comparable, rather than each being computed on its own sample-specific range. From each PI we extract seven summary statistics, described below.

The \emph{total mass}, defined as the sum of all pixel intensities (equivalently, the integral of $\rho$ over the sampled birth--persistence region), 
measures the aggregate weighted intensity captured within the sampled birth--persistence region; it is large when the diagram contains features with both long lifetimes and non-negligible weight $w(\ell)$.

The \emph{max pixel} is the highest pixel intensity in the PI, indicating a region of highly concentrated persistence intensity. The fraction of pixels whose intensities exceed the average intensity of their own PI is collected as the \emph{active fraction}. 
This statistic records the fraction of the image with intensity above its own mean value.

Viewing the PI as a mass distribution, the \emph{birth centroid} and \emph{persistence centroid} are its centers of mass along the birth and persistence axes, respectively. Denoting the pixel intensities as $\mathrm{PI}_{m,n}$, with $m$ and $n$ indexing the birth and persistence axes, respectively, the two centroids are given by
\begin{align}
    \overline{b}
    &= \frac{\sum_{m,n}\mathrm{PI}_{m,n}\,x_m}{M},\\
    \overline{\ell}
    &= \frac{\sum_{m,n}\mathrm{PI}_{m,n}\,y_n}{M},
\end{align}
where $M=\sum_{m,n}\mathrm{PI}_{m,n}$, and $x_m$ and $y_n$ denote the normalized birth and persistence coordinates associated with pixel row $m$ and column $n$, respectively. The two centroids indicate where the PI mass is concentrated along the birth and persistence axes.

For $M>0$, the probability mass $q_{m,n}=\mathrm{PI}_{m,n}/M$ can further be used to compute the normalized \emph{entropy},
\begin{align}
    E=
    \frac{-\sum_{m,n}q_{m,n}\log q_{m,n}}
    {\log\!\left(n_{\rm pi}^2\right)},
\end{align}
with the convention $0\log 0=0$. Since $n_{\rm pi}^2$ is the total number of pixels, $E\in[0,1]$. The entropy measures how spread out or concentrated the normalized pixel intensities are across the two-dimensional image, with $E=0$ when all mass is concentrated in a single pixel and $E=1$ for a uniform distribution over all pixels.

Finally, the \emph{90th-percentile intensity} is the pixel intensity value below which $90\%$ of pixel intensities fall. Unlike the maximum intensity, it is not determined by a single extreme pixel and therefore provides a less outlier-sensitive summary of the upper end of the pixel-intensity distribution.

When the PI has zero total mass ($M=0$), as occurs for an empty persistence diagram, the mass-normalized quantities and centroids are mathematically undefined. In this case, we define all seven PI summary statistics to be zero by convention, including $E=0$.

When the PI has zero total mass ($M=0$), as occurs for an empty persistence diagram, the mass-normalized quantities and centroids are mathematically undefined. In this case, we define all seven PI summary statistics to be zero by convention, including $E=0$.

An empty persistence diagram at a given homology dimension, which can occur when no topological features are detected, is handled by convention in the other three branches as well: the entropy is defined as $H_{\rm ent}=0$, all five lifetime-branch statistics are defined as $0$, and the Betti curve is defined as identically zero across all sampled thresholds, so that the resulting maximum, mean, standard deviation, peak position, and turning-point count are each $0$. The bimodality coefficient is not covered by this convention: since it is computed as a ratio with a numerically stabilized denominator, an identically zero Betti curve instead yields a large finite value rather than $0$.

\subsubsection{Learnable fusion and classification}\label{sec:alg-training}

So far, we have extracted five feature groups from the geometric, entropy, lifetime, Betti, and PI branches, all computed deterministically. Since these feature groups arise from different geometric and topological constructions, we give each group its own batch-normalization module and its own projection into a shared embedding space of dimension $D_{\rm emb}$, rather than sharing a single normalization-and-projection module across all groups. These two steps are carried out by the \emph{group projector}, yielding five $D_{\rm emb}$-dimensional embeddings, or \emph{tokens}, one per feature group.

The five tokens are then combined by a \emph{fusion layer}, which allows each group to access information from every other group before classification. We consider five interchangeable fusion strategies, differing in their number of parameters and expressivity, each mapping the five $D_{\rm emb}$-dimensional tokens $\{\mathrm{token}_1,\ldots,\mathrm{token}_5\}$ to a single fused representation.

\paragraph{Bilinear pairwise fusion.}
For every pair of groups $(i,j)$ with $i<j$, a scalar interaction
\begin{align}
s_{ij}
=
\tanh\!\left(
\sum_{m=1}^{D_{\rm emb}}
(\mathrm{token}_i)_m
(w_{ij})_m
(\mathrm{token}_j)_m
\right),
\end{align}
is computed from learned weights $w_{ij}\in\mathbb{R}^{D_{\rm emb}}$ and converted into a shared residual $\delta_{ij}=s_{ij}v_{ij}$, where $v_{ij}\in\mathbb{R}^{D_{\rm emb}}$ is a second learned vector, added back to both token $i$ and token $j$. Once all pairwise residuals have been accumulated, each token is passed through its own LayerNorm before the resulting tokens are concatenated into the fused representation. This is the cheapest strategy, capturing direct pairwise group interactions without attention or positional structure.

\paragraph{Gated residual fusion.}
For each token $i$, the mean of all other tokens,
\begin{align}
{\rm context}_{i}
=
{\rm mean}\!\left(
\{{\rm token}_{j}:j\neq i\}
\right),
\end{align}
is processed through a learned sigmoid gate:
\begin{align}
\mathrm{gate}_i
&=
\sigma(W_{g,i}\,\mathrm{context}_i),\\
\mathrm{value}_i
&=
\tanh(W_{v,i}\,\mathrm{context}_i),
\end{align}
with $W_{g,i},W_{v,i}\in\mathbb{R}^{D_{\rm emb}\times D_{\rm emb}}$ learned per-token weight matrices. The token is then updated as
\begin{align}
{\rm LayerNorm}
\left(
\mathrm{token}_i
+
\mathrm{gate}_i\odot\mathrm{value}_i
\right).
\end{align}
When the gate approaches zero, the cross-group residual contribution vanishes and the update approaches ${\rm LayerNorm}(\mathrm{token}_i)$.
The five updated tokens are then concatenated into the fused representation.

\paragraph{Linear attention fusion.}
Query--key--value attention among the five tokens is used, with the softmax replaced by the positive feature map
\begin{align}
\phi(x)=\mathrm{ELU}(x)+1,
\end{align}
following the linear-attention formulation of Katharopoulos \textit{et al.}~\cite{Katharopoulos2020}.
Here, ${\rm ELU}$ denotes the exponential linear unit~\footnote{The exponential linear unit reads
$\displaystyle
{\rm ELU}(x)=x
$
for $x>0$ and
$\displaystyle
{\rm ELU}(x)=e^x-1
$
for $x\leq0$.
}
The feature map $\phi$ is applied independently to each of $n_{\rm head}$ attention heads, where $n_{\rm head}$ is chosen such that $D_{\rm emb}$ is divisible by $n_{\rm head}$.

For each query token $q_i$, linear attention is computed per head as
\begin{align}
{\rm Attn}(q_i)
=
\frac{
\phi(q_i)^\top
\left(
\sum_{j=1}^{N_{\rm t}}
\phi(k_j)v_j^\top
\right)
}{
\phi(q_i)^\top
\left(
\sum_{j=1}^{N_{\rm t}}
\phi(k_j)
\right)
},
\end{align}
rather than using softmax attention.

Here, $q_i,k_i,v_i\in\mathbb{R}^{d_{\rm h}}$ are the per-head query, key, and value vectors, $N_{\rm t}=5$ is the number of tokens, and $d_{\rm h}=D_{\rm emb}/n_{\rm head}$ is the per-head dimension. Each layer applies a pre-norm residual attention block, followed by an output projection and a second pre-norm residual feedforward sublayer. After the final layer, the five tokens are concatenated into the fused representation.

\paragraph{Low-rank cross-group MLP.}
All five tokens are concatenated into $x\in\mathbb{R}^{5D_{\rm emb}}$. This vector is then passed through $n_{\rm layers}$ stacked bottleneck blocks, each with bottleneck dimension $n_{\rm rank}$ and a residual connection. For $l=1,\ldots,n_{\rm layers}$, each block computes
\begin{align}
    h^{(l)}
    &=
    \mathrm{GELU}\!\left(
    W_{\mathrm{down}}^{(l)}x^{(l-1)}
    \right),
\end{align}
with
$W_{\mathrm{down}}^{(l)}\in\mathbb{R}^{n_{\rm rank}\times 5D_{\rm emb}}$, followed by
\begin{align}
    x^{(l)}
    &=
    \mathrm{LayerNorm}\!\left(
    x^{(l-1)}+
    W_{\mathrm{up}}^{(l)}h^{(l)}
    \right),
\end{align}
with
$W_{\mathrm{up}}^{(l)}\in\mathbb{R}^{5D_{\rm emb}\times n_{\rm rank}}$.
Here, $x^{(0)}=x$ is the concatenated token vector and $x^{(n_{\rm layers})}$ is the final fused representation.
Because the bottleneck layers act on the concatenated representation, they allow information from all five feature groups to be mixed while limiting the number of trainable parameters through the dimension $n_{\rm rank}$. 
Because the bottleneck layers act on the concatenated representation, they allow information from all five feature groups to be mixed while limiting the number of trainable parameters through the dimension $n_{\rm rank}$, and this fusion strategy serves as \TopTimeNet{}'s default.

\paragraph{Multi-group token attention.}
The five tokens $\{{\rm token}_1,\ldots,{\rm token}_5\}$ are stacked into
$X\in\mathbb{R}^{5\times D_{\rm emb}}$ and offset by a learned positional embedding
$P\in\mathbb{R}^{5\times D_{\rm emb}}$, so that $X^{(0)}=X+P$ distinguishes group identity independently of token content, since each of the five positions consistently corresponds to the same feature group across all inputs.

The result is passed through a standard pre-norm Transformer encoder with $n_{\rm layers}$ layers, each applying multi-head self-attention (with head count as a configurable hyperparameter, subject to $D_{\rm emb}$ divisibility) over the five tokens, followed by an output projection and a feedforward sublayer, both with pre-norm residual connections. This is the same block structure described for linear attention fusion, but using standard softmax attention rather than the kernelized formulation. The five output tokens are then concatenated into the fused representation.
This strategy introduces standard softmax self-attention across the five feature-group tokens and generally uses more trainable parameters than the simpler fusion schemes considered above.

All five strategies share a common interface, so the choice of fusion strategy is treated as a hyperparameter.
The fused representation is passed to a \emph{classification head}: a single batch-normalization layer applied to the full fused representation, followed by a multilayer perceptron with a configurable number of hidden layers, each followed by an activation function and dropout, which outputs the final class logits.
When no hidden layers are used, the batch-normalized fused representation is mapped directly to the class logits by a single linear layer.

\subsection{Model training and hyperparameter selection}\label{sec:hyp-selection}

\subsubsection{Training protocol}\label{sec:training-protocol}

The learnable components of \TopTimeNet{} are trained jointly with a cross-entropy loss. 
The optimizer is selected between Adam~\cite{kingma2014adam} and AdamW~\cite{loshchilov2017adamw}.
The optimizer, label smoothing, gradient-clipping norm, and the input-normalization scheme described below are hyperparameters selected independently for each configuration by the random search of Sec.~\ref{sec:hyp-search}; the small and large configurations compared in Sec.~\ref{sec:results} therefore differ in some of these settings, as detailed in Table~\ref{tab:model-comparison}. Training runs for up to $500$ epochs, with 
early stopping using a patience of $50$ epochs and restoration of the best validation checkpoint,
and the learning rate is reduced on validation-loss plateaus via \texttt{ReduceLROnPlateau}.

Data are split $80/10/10$ into training, validation, and test sets, stratified by class. This split is performed at the level of individual segments rather than source trajectories, using stratified random shuffling. As a result, segments originating from the same simulated trajectory may appear in more than one partition, 
which may inflate the reported test performance relative to generalization on entirely unseen trajectories. This protocol also does not assess generalization to dynamical systems excluded entirely from training, which would require a system-level split.
However, all benchmarked architectures, including \TopTimeNet{} and every baseline reported in Sec.~\ref{sec:results}, are trained and evaluated on identical segment-level splits under this same procedure; the use of identical splits nevertheless enables a controlled comparison between architectures under the same segment-level evaluation protocol. 
Trajectory-level and system-level splitting are therefore important directions for future work when the goal is to estimate generalization to unseen trajectories or unseen dynamical systems, respectively.

The raw time series are not rescaled before the geometric and topological features are computed; each simulated system's signal retains its native amplitude, so that diameter, nearest-neighbor distances, and other scale-dependent quantities reflect the native scale of the underlying simulated variable. An optional normalization stage, selected by the hyperparameter search, may standardize the resulting $42$-dimensional feature vector using training-set statistics (mean and standard deviation per feature dimension) before it is passed to the model; this is independent of, and precedes, the per-group batch normalization already applied inside the group projector (Sec.~\ref{sec:alg-training}). Of the two configurations reported in Table~\ref{tab:model-comparison}, the small configuration applies this per-feature standardization, while the large configuration does not.

After training, a single calibration temperature $T$ is fit on the validation set via LBFGS to minimize the cross-entropy of temperature-scaled logits, with $T$ constrained to $[0.5, 5.0]$ during optimization. This aims to improve the alignment between predicted confidence and empirical accuracy without altering the trained weights~\cite{guo2017calibration}.

\subsubsection{Hyperparameter search}\label{sec:hyp-search}

Hyperparameters are selected via random search over $400$ independently sampled configurations, spanning architectural choices (embedding dimension, fusion strategy, fusion rank, number of attention layers/heads, classifier head width, activation, dropout), regularization strength (label smoothing, gradient clipping norm), and optimization settings (learning rate, optimizer, input normalization scheme). Batch size and early-stopping patience are both nominally part of the search space but restricted to a single possible value ($128$ and $50$ epochs, respectively) for every trial, and so are, in effect, fixed throughout the search and final evaluation rather than genuinely varied. Sampled configurations that pair an attention-based fusion strategy (linear attention or the multi-group transformer) with an embedding dimension not divisible by the number of attention heads are rejected and resampled, since multi-head attention requires this divisibility. TDA-specific hyperparameters (Takens embedding dimension and delay, number of Betti bins, persistence-image resolution and kernel width) are fixed at the values described in Sec.~\ref{sec:alg-features} rather than included in the search, and the corresponding TDA features are precomputed and cached once for all trials; since these parameters do not vary across trials, a single cached feature set is valid throughout the search. The input normalization scheme is included as a tunable hyperparameter during the search; the two selected configurations identified below differ in this setting (Sec.~\ref{sec:training-protocol}, Table~\ref{tab:model-comparison}). To keep the search computationally tractable, each trial trains a single model instance for up to $500$ epochs, rather than the multiple independent runs used for final evaluation (Sec.~\ref{sec:training-protocol}), and both post-hoc temperature scaling and the noise-robustness sweep are turned off during search, as neither affects which configuration is selected.

Each trial is scored by a weighted combination of macro F1 and the geometric mean of sensitivity and specificity, $\text{score}=\alpha\cdot F1+(1-\alpha)\cdot \text{G-mean}$, with $\alpha=0.6$, chosen to balance macro-F1 performance with sensitivity--specificity balance. The CNN and Transformer baselines of Sec.~\ref{sec:baselines} were selected using the same scoring formula and the same random-search, validity-rejection, and Pareto-selection procedures described in this subsection, but over their own architecture-specific search spaces, with $250$ trials each and $\alpha=0.7$; we note this difference in objective weight explicitly, as it was not matched to \TopTimeNet{}'s own search. Across all $400$ trials, we identify the configuration achieving the highest score and select it as the best-performing (``large'') set of hyperparameters for \TopTimeNet{}. We additionally identify a parameter-efficient (``small'') alternative via a Pareto-dominance search on the same (score, parameter-count) trials, using a score tolerance of $1\%$: a configuration is retained on the Pareto front only if no other trial simultaneously scores within $1\%$ of it while using strictly fewer parameters, and among the retained configurations we report the highest-scoring one as the parameter-efficient candidate. 
The $1\%$ tolerance allows configurations with similar predictive performance to be compared on parameter count, thereby favoring smaller models when the reduction in score is limited.
The selected small and large configurations are each retrained $30$ times with independent random seeds, and we report the mean $\pm$ standard deviation of the evaluation metrics across these runs.

\section{Illustrative example}\label{sec:example}

Before presenting the results of \TopTimeNet{}, we illustrate how the proposed features help distinguish periodic from chaotic dynamics using the time evolution of a double pendulum.

The double pendulum is a classical system in mechanics. Its time evolution is governed by
\begin{widetext}
\begin{align}
    \dot{\theta_1} &= \omega_1 ,\\
    \dot{\theta_2} &= \omega_2 ,\\
    \dot{\omega_1} &=
    \frac{-g (2 m_1+m_2) \sin(\theta_1) -m_2 g \sin(\theta_1 -2\theta_2) -2 \sin(\theta_1 - \theta_2) m_2 (\omega_2^2 l_2 + \omega_1^2 l_1 \cos(\theta_1 - \theta_2))}{l_1 (2m_1 + m_2 - m_2 \cos(2 \theta_1 -2\theta_2))}
    ,\\
    \dot{\omega_2} &=
    \frac{2 \sin(\theta_1-\theta_2)( \omega_1^2 l_1 (m_1+m_2) + g(m_1+m_2) \cos(\theta_1) + \omega_2^2 l_2 m_2 \cos(\theta_1 -\theta_2))}{ l_2 (2m_1 +m_2 -m_2 \cos(2\theta_1 -2 \theta_2))}
    .
\end{align}
\end{widetext}

Here, the parameters are $g=9.81\,{\rm m/s^2}$, $m_1=m_2=1\,{\rm kg}$, and $l_1=l_2=1\,{\rm m}$. This system exhibits a periodic response for initial conditions $\theta_1(0)=0.4\,{\rm rad}$, $\theta_2(0)=0.6\,{\rm rad}$, $\omega_1(0)=\omega_2(0)=1.0\,{\rm rad/s}$, and a chaotic response for $\theta_1(0)=0.0\,{\rm rad}$, $\theta_2(0)=3.0\,{\rm rad}$, $\omega_1(0)=\omega_2(0)=0.0\,{\rm rad/s}$, using the parameter presets of the Teaspoon library~\cite{Khasawneh2025}. 

These two regimes were further characterized by the largest Lyapunov exponent, estimated using the method of Eckmann \textit{et al.}~\cite{Eckmann1986}: 
$\lambda_1 \approx 0.015$ for the periodic trajectory and 
$\lambda_1 \approx 0.106$ for the chaotic trajectory. The former value is close to zero, consistent with regular motion up to numerical and finite-time estimation effects, whereas the latter is clearly positive and indicates sensitive dependence on initial conditions.

\begin{figure}[ht!]
    \centering
    \includegraphics[width=0.99\linewidth]{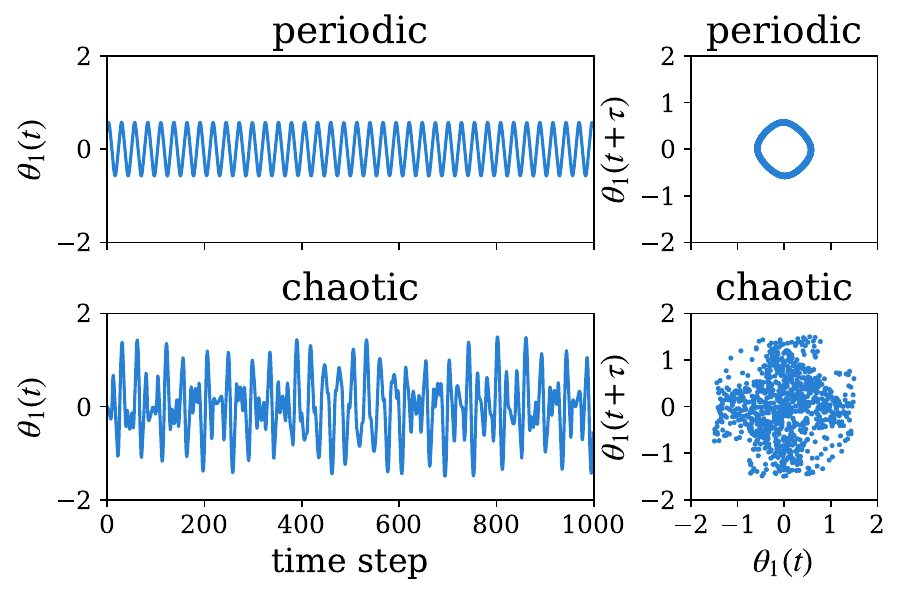}
    \caption{\textbf{Time-dependent $\theta_1$ and its Takens embedding.} Left panels present $\theta_1(t)$ for the periodic (top) and chaotic (bottom) regimes. Right panels show the associated point clouds obtained via a Takens embedding with time delay $\tau=20$ and embedding dimension $d_{\rm emb}=2$.}
    \label{fig:dp_phase_space}
\end{figure}

The time evolution of $\theta_1$ is shown in the left panels of Fig.~\ref{fig:dp_phase_space}, and the corresponding point cloud for each regime, obtained using the Takens embedding described in Sec.~\ref{sec:alg-features}, is shown in the right panels.

The contrast between the two regimes is already visible at the level of the raw signal and its embedding. In the periodic case, $\theta_1(t)$ shows regular, repeating oscillations, and its Takens embedding forms a clean, closed one-dimensional loop in phase space, consistent with a periodic orbit. In the chaotic case, $\theta_1(t)$ shows irregular oscillations of varying amplitude, and the corresponding point cloud instead fills a broad, diffuse region of the two-dimensional embedding space with no discernible loop structure, reflecting the absence of any single dominant periodic orbit.

\begin{table}[ht!]
    \centering
    \caption{Geometric branch feature values for the double pendulum, periodic vs.\ chaotic regime.}
    \label{tab:dp_geometric}
    \begin{tabular}{lcc}
        \hline
        \textbf{Feature} & \textbf{Periodic} & \textbf{Chaotic} \\
        \hline
        Diameter                    & $1.1552$ & $3.2792$ \\
        Correlation-dimension proxy & $1.0672$ & $1.6585$ \\
        Mean nearest-neighbor dist. & $0.0014$ & $0.0410$ \\
        Std.\ nearest-neighbor dist.& $0.0005$ & $0.0293$ \\
        \hline
    \end{tabular}
\end{table}

The four features computed by the geometric branch of \TopTimeNet{} for each signal are given in Table~\ref{tab:dp_geometric}, and they capture this contrast quantitatively. The diameter nearly triples between the two regimes ($1.1552 \to 3.2792$), and the mean nearest-neighbor distance increases by a factor of approximately $29$ and its standard deviation by a factor of approximately $59$, consistent with the broader spatial spread of the chaotic point cloud. The correlation-dimension proxy increases as well, from $1.0672$ to $1.6585$, consistent with the chaotic point cloud having a higher effective dimension than the near-one-dimensional periodic trajectory. Together, these four features already separate the two regimes in this illustrative example using only the geometry of the point cloud, before any topological information is introduced; 
the topological branches discussed next provide complementary descriptors of the structure of these point clouds.

\begin{figure}[ht!]
    \centering
    \includegraphics[width=0.95\linewidth]{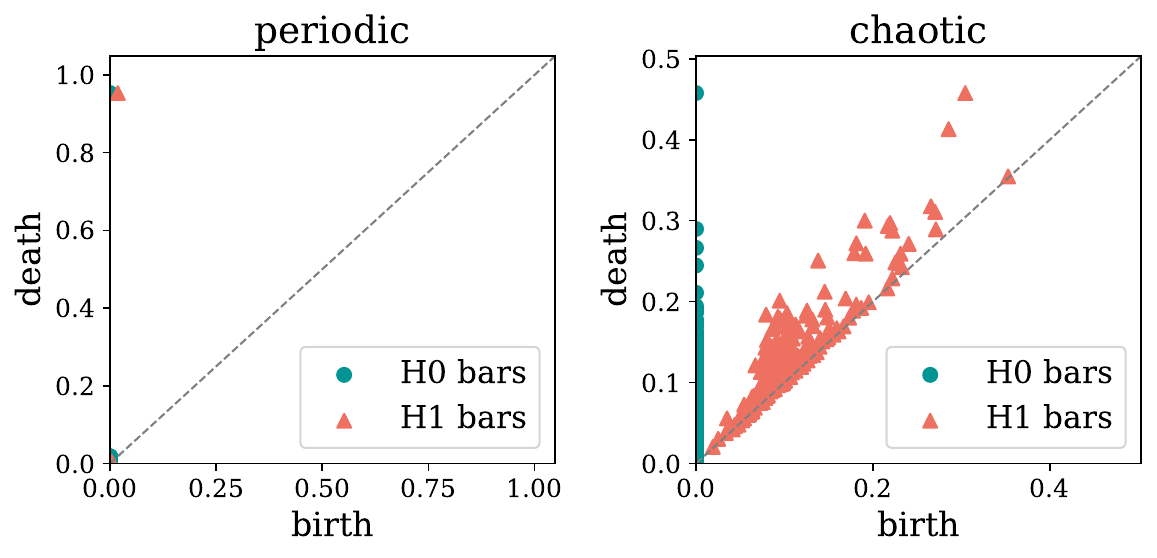}
    \caption{\textbf{Persistence diagrams for the two tracked homology dimensions.} $H_0$ (connected components, circles) and $H_1$ (loops, triangles) features are shown for the periodic (left) and chaotic (right) point clouds of Fig.~\ref{fig:dp_phase_space}; note the different axis ranges between panels. Points farther from the diagonal correspond to more persistent topological features.}
    \label{fig:dp_persistence}
\end{figure}

Figure~\ref{fig:dp_persistence} shows the resulting persistence diagrams~\footnote{The single essential $H_{0}$ class, whose death time is formally infinite, is capped at the largest finite death value observed 
across all tracked homology dimensions for that point cloud before any downstream statistics are computed; this ensures every reported quantity, including the entropy, lifetime, Betti-curve, and persistence-image features, is finite-valued by construction.}. In the periodic case, the $H_1$ diagram is dominated by a single point far from the diagonal, at a death value close to $1$, associated with a single, highly persistent loop, consistent with the clean closed orbit seen in the point cloud. In the chaotic case, the $H_1$ diagram instead contains hundreds of points clustered close to the diagonal, with only a handful reaching moderately large death values; this reflects a large number of short-lived loops rather than one dominant structure, and is captured by the entropy branch (Sec.~\ref{sec:alg-features}), as reflected in Table~\ref{tab:dp_entropy}. The $H_0$ entropy values are less strongly separated between the two regimes, since both point clouds start with the same number of connected components and contain many short-lived finite $H_0$ classes (Fig.~\ref{fig:dp_persistence}), so $H_0$ entropy primarily reflects the distribution of component-merging scales in this example. The $H_1$ entropy, by contrast, shows a far more striking difference. Here, the periodic case has a near-zero value ($0.038$), consistent with the single dominant loop, whereas the value is $5.01$ for the chaotic case, reflecting the many scattered points of comparable lifetime seen in the diagram above and yielding a much stronger contrast than for $H_0$ entropy.

\begin{table}[ht!]
    \centering
    \caption{Entropy branch feature values for the double pendulum, periodic vs.\ chaotic regime.}
    \label{tab:dp_entropy}
    \begin{tabular}{lcc}
        \hline
        \textbf{Feature} & \textbf{Periodic} & \textbf{Chaotic} \\
        \hline
        Entropy, $H_0$ & $5.5456$ & $6.6991$ \\
        Entropy, $H_1$ & $0.0379$ & $5.0058$ \\
        \hline
    \end{tabular}
\end{table}

The third branch with five features is the lifetime branch, whose values for the periodic and chaotic signals are given in Table~\ref{tab:dp_lifetime}. Unlike entropy, which reduces the full lifetime distribution to a single number, these five statistics decompose it along complementary axes, namely, overall scale (maximum and total lifetime), concentration (dominance ratio), spread (coefficient of variation), and count (number of significant lifetimes).

\begin{table*}[ht!]
    \centering
    \caption{Lifetime branch feature values for the double pendulum, periodic vs.\ chaotic regime.}
    \label{tab:dp_lifetime}
    \setlength{\tabcolsep}{12pt}
    \begin{tabular}{lcccc}
        \hline
        & \multicolumn{2}{c}{\textbf{$H_0$}} & \multicolumn{2}{c}{\textbf{$H_1$}} \\
        \textbf{Feature} & \textbf{Periodic} & \textbf{Chaotic} & \textbf{Periodic} & \textbf{Chaotic} \\
        \hline
        Maximum lifetime        & $0.9532$ & $0.4576$  & $0.9348$ & $0.1533$ \\
        Total lifetime          & $4.8949$ & $52.1417$ & $0.9386$ & $6.0242$ \\
        Dominance ratio         & $0.1947$ & $0.0088$  & $0.9959$ & $0.0254$ \\
        Coefficient of variation& $6.1556$ & $0.6769$  & $5.0775$ & $1.0785$ \\
        Number of significant lifetimes & $1$ & $836$  & $1$      & $169$   \\
        \hline
    \end{tabular}
\end{table*}

The $H_0$ statistics tell a comparatively modest story. Neither regime is dominated by a single $H_0$ lifetime, as indicated by the dominance ratios of $0.1947$ and $0.0088$, consistent with the dense sampling of nearby trajectory points discussed for the entropy branch; the chaotic case nonetheless still stands out through its far larger total lifetime ($52.1417$ vs.\ $4.8949$) and number of significant lifetimes ($836$ vs.\ $1$), consistent with a broader spatial spread of the point cloud rather than any single dominant merging event.

At $H_1$, by contrast, the distinction between regimes is stark and consistent across every feature. The periodic signal's dominance ratio is $0.9959$ and its number of significant lifetimes is exactly $1$, indicating that nearly all of the persistence is carried by a single loop, with essentially nothing left over for any other feature. In the chaotic signal, the most persistent loop
carries only $2.5\%$ of the total persistence (dominance ratio $0.0254$), with persistence distributed across $169$ significant bars, suggesting contributions from many additional features. The total lifetime grows more than sixfold ($0.9386 \to 6.0242$) despite the maximum lifetime of any single bar actually \emph{shrinking} ($0.9348 \to 0.1533$). A larger total lifetime together with a smaller maximum lifetime requires substantial contributions from additional features rather than concentration in a single one; the coefficient of variation further indicates a less dispersed relative distribution of lifetimes, as it drops from $5.0775$ in the periodic case to $1.0785$ in the chaotic case. 
This is consistent with the entropy and dominance-ratio results: the periodic $H_1$ persistence is dominated by a single feature, whereas the chaotic case distributes persistence across many features.

Together, the five lifetime statistics, particularly evaluated at $H_1$, recover the same periodic/chaotic distinction as the entropy branch, but decomposed into interpretable components (how big, how concentrated, how spread out, how many) rather than a single aggregate number, illustrating the complementary role the lifetime branch plays alongside entropy in \TopTimeNet{}'s topological feature set.

\begin{figure}[ht!]
    \centering
    \includegraphics[width=0.99\linewidth]{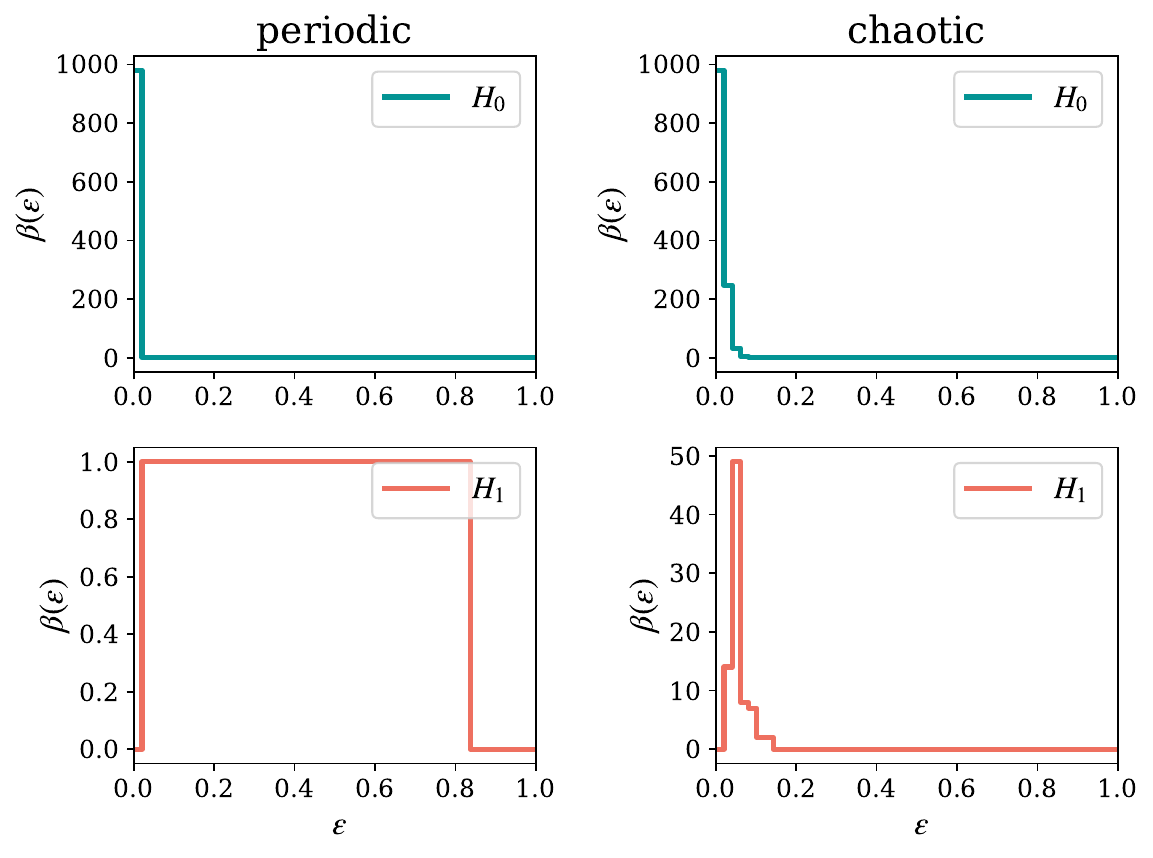}
    \caption{\textbf{Diameter-normalized Betti curves $\beta(\varepsilon)$ for $H_0$ (top) and $H_1$ (bottom), periodic (left) and chaotic (right).} Note the different vertical scales between panels.}
    \label{fig:dp_betti}
\end{figure}

Figure~\ref{fig:dp_betti} presents the corresponding Betti curves. For $H_0$, both regimes show a sharp initial spike as the $980$ point-cloud points merge into progressively fewer connected components as the filtration threshold grows, but the chaotic curve decays more gradually, with an intermediate plateau 
around a normalized filtration threshold $\varepsilon$ of $0.05$--$0.1$,
reflecting a less uniform spatial distribution of points than in the periodic case. The $H_1$ curves show a far more striking contrast. The periodic curve is a single clean rectangular plateau at $\beta=1$, persisting across most of the normalized filtration range, indicating a single loop that remains alive for nearly the entire filtration sweep. The chaotic curve instead rises sharply to a peak of nearly $50$ simultaneously coexisting loops 
at small normalized filtration thresholds $\varepsilon$,
before decaying through several irregular steps to near zero 
by a normalized filtration threshold of approximately $0.15$,
reflecting many loops appearing and rapidly disappearing early in the filtration.

\begin{figure}[ht!]
    \centering
    \includegraphics[width=0.99\linewidth]{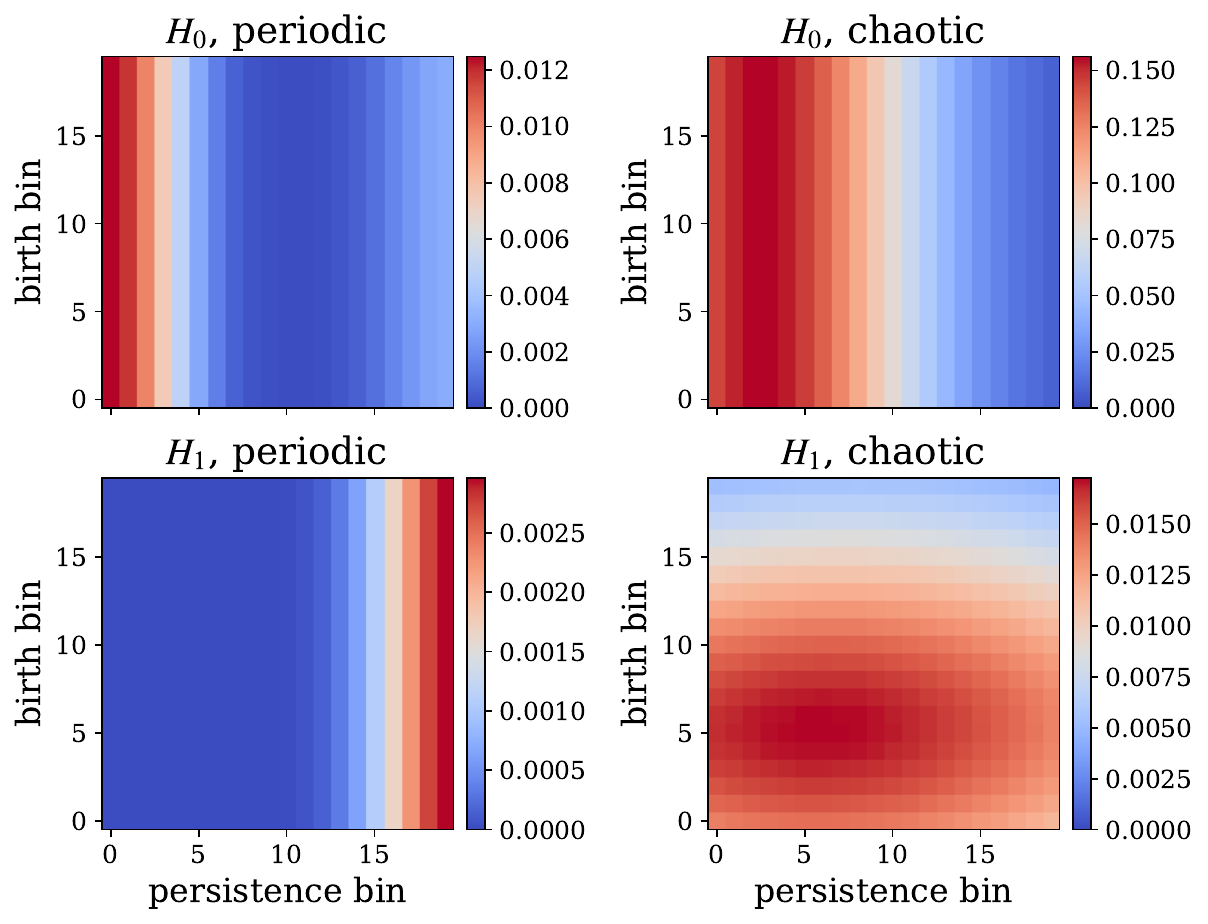}
    \caption{\textbf{Persistence images for $H_0$ (top) and $H_1$ (bottom), periodic (left) and chaotic (right).} Color indicates pixel intensity (note the different color scales between panels); axes are birth and persistence pixel bins.}
    \label{fig:dp_pi}
\end{figure}

The six summary statistics extracted from these curves, given in Table~\ref{tab:dp_betti}, largely mirror this picture. At $H_0$, the two regimes are nearly indistinguishable. The maximum, peak position, and number of detected turning points are identical between periodic and chaotic, and the mean and standard deviation differ only modestly ($20.40$ vs.\ $25.38$ and $137.09$ vs.\ $140.74$, respectively), 
consistent with both point clouds containing the same number of densely sampled trajectory points and many short-lived connected components.
The strongest differences instead occur at $H_1$, where most statistics change substantially between regimes. The maximum jumps from a single coexisting loop ($1$) to nearly fifty ($49$), and the standard deviation grows by more than a factor of $17$ ($0.40 \to 7.19$). One turning point is detected in the chaotic curve under our adopted criterion, which counts sign changes between immediately adjacent nonzero first differences $\Delta \beta_j$~\footnote{Because Betti curves are piecewise constant, this criterion does not count turning points separated by one or more zero-valued differences (plateaus); it therefore provides a conservative lower bound on the number of true direction changes in the curve, rather than an exhaustive count.}, whereas none is detected in the periodic curve under the same criterion. 

The bimodality coefficient, computed from the statistical distribution of the sampled curve values 
$\beta_0,\ldots,\beta_{n_{\rm bin}-1}$
rather than from the number of peaks of $\beta(\varepsilon)$ as a function of filtration scale, changes only mildly between regimes (from $1.00$ to $0.92$); 
as discussed above, this coefficient is a skewness--kurtosis-based heuristic and should not be interpreted as a direct count or test of peaks in the Betti curve.
A more direct picture of where these loops are concentrated along the filtration range is instead given by the peak position, which shifts from $0.0204$ to $0.0408$ between regimes.

\begin{table*}[ht!]
    \centering
    \caption{Betti curve branch feature values for the double pendulum, periodic vs.\ chaotic regime.}
    \label{tab:dp_betti}
    \setlength{\tabcolsep}{12pt}
    \begin{tabular}{lcccc}
        \hline
        & \multicolumn{2}{c}{\textbf{$H_0$}} & \multicolumn{2}{c}{\textbf{$H_1$}} \\
        \textbf{Feature} & \textbf{Periodic} & \textbf{Chaotic} & \textbf{Periodic} & \textbf{Chaotic} \\
        \hline
        Max                & $980.0000$ & $980.0000$ & $1.0000$ & $49.0000$ \\
        Mean                & $20.4000$  & $25.3800$  & $0.8000$ & $1.6400$  \\
        Standard deviation  & $137.0863$ & $140.7353$ & $0.4000$ & $7.1882$  \\
        Peak position       & $0.0000$   & $0.0000$   & $0.0204$ & $0.0408$  \\
        Turning points & $0$        & $0$        & $0$      & $1$       \\
        Bimodality coefficient & $0.9999$ & $0.9626$  & $1.0000$ & $0.9247$  \\
        \hline
    \end{tabular}
\end{table*}

Finally, Fig.~\ref{fig:dp_pi} shows the persistence images. The $H_0$ images are qualitatively similar between regimes. Both show intensity concentrated at low persistence values near the common $H_0$ birth location, since most components merge almost immediately, but the chaotic image's peak intensity is roughly an order of magnitude higher ($0.0125 \to 0.1561$), reflecting differences in the persistence distribution and the resulting weighted density. The $H_1$ images show the clearest visual distinction. The periodic image shows its mass concentrated toward large persistence values (persistence centroid $0.89$) at its corresponding birth location, mirroring the single long-lived loop seen in the persistence diagram and Betti curve.
The chaotic image instead shows a visually diffuse region of intensity centered at small-to-moderate birth and persistence values (persistence centroid $0.49$), with intensity spread more broadly across the grid. Its active fraction is $0.60$, compared with $0.30$ for the periodic image.

\begin{table*}[ht!]
    \centering
    \caption{Persistence image branch feature values for the double pendulum, periodic vs.\ chaotic regime.}
    \label{tab:dp_pi}
    \setlength{\tabcolsep}{12pt}
    \begin{tabular}{lcccc}
        \hline
        & \multicolumn{2}{c}{\textbf{$H_0$}} & \multicolumn{2}{c}{\textbf{$H_1$}} \\
        \textbf{Feature} & \textbf{Periodic} & \textbf{Chaotic} & \textbf{Periodic} & \textbf{Chaotic} \\
        \hline
        Total mass                & $1.2801$ & $34.7081$ & $0.2390$ & $5.0567$ \\
        Max pixel                 & $0.0125$ & $0.1561$  & $0.0030$ & $0.0173$ \\
        Active fraction           & $0.2500$ & $0.5000$  & $0.3000$ & $0.6000$ \\
        Birth centroid            & $0.5000$ & $0.5000$  & $0.5000$ & $0.4265$ \\
        Persistence centroid      & $0.2542$ & $0.3124$  & $0.8928$ & $0.4860$ \\
        90th-percentile intensity & $0.0102$ & $0.1531$  & $0.0023$ & $0.0167$ \\
        Entropy                   & $0.8895$ & $0.9610$  & $0.8127$ & $0.9925$ \\
        \hline
    \end{tabular}
\end{table*}

The seven summary statistics extracted from these images, given in Table~\ref{tab:dp_pi}, are consistent with this picture. At $H_0$, the total mass and max pixel both increase sharply from periodic to chaotic ($1.2801 \to 34.7081$ and $0.0125 \to 0.1561$, respectively), reflecting the greater aggregate weighted intensity
and higher peak intensity of the chaotic image; the active fraction also increases from $0.25$ to $0.50$.

The birth centroid is fixed at exactly $0.5$ in both regimes; 
this is a consequence of how the degenerate $H_0$ birth axis is handled in our implementation rather than a property of the underlying dynamics.
Because every $H_0$ feature is born at the start of the Vietoris--Rips filtration, all $H_0$ birth values are identical, and the fitted birth axis collapses to zero width. Our implementation pads this degenerate axis to a single pixel before resizing to the target grid resolution; the resulting interpolation distributes the single row of birth-axis mass uniformly across all rows of the resized image, which places the birth centroid exactly at the midpoint of the normalized $[0,1]$ range by symmetry, regardless of regime. The persistence centroid, by contrast, shifts modestly with regime ($0.2542 \to 0.3124$), since it is computed along the (non-degenerate) persistence axis.

At $H_1$, the persistence centroid shows a clear contrast between the two regimes:
it sits at $0.8928$ for the periodic case, consistent with the mass being concentrated toward large persistence values as seen in the image, versus $0.4860$ for the chaotic case, where mass is instead spread toward more moderate persistence values. The birth centroid also shifts noticeably at $H_1$ ($0.5000 \to 0.4265$), where, unlike $H_0$, birth times genuinely vary across features, so this shift reflects the chaotic image's mass being centered at somewhat earlier birth times than the periodic one. 
Entropy increases from $0.8127$ to $0.9925$, consistent with the more diffuse intensity distribution visible in the chaotic image compared with the narrower band in the periodic case. The active fraction also increases from $0.3000$ to $0.6000$.

Taken together, this illustrative example shows how the five branches provide complementary descriptions of the contrast between periodic and chaotic dynamics.
The geometric branch shows that the chaotic point cloud occupies a larger, less uniformly sampled region of phase space; the entropy branch shows that its topological features are less dominated by a single persistent structure; the lifetime branch decomposes this same distinction into interpretable measures of scale, concentration, and count; the Betti curve branch reveals the same contrast unfolding across the filtration itself, from a single stable plateau to a sharp, jagged, rapidly decaying peak; and the persistence image branch localizes where in the birth--persistence plane this difference occurs.

In this example, the $H_1$ features show the clearest topological contrast between the two regimes,
although several $H_0$ features also differ substantially, highlighting the complementary information provided by connectivity- and loop-level topology when characterizing periodic versus chaotic dynamics.

Several individual features already distinguish the two trajectories in this illustrative example, but the fact that the distinction appears across geometry, entropy, lifetime statistics, Betti curves, and persistence images motivates combining all $42$ features rather than relying on any single representation. Appendix~\ref{app:raw-signal-illustration} extends this same double-pendulum example to illustrate how raw-signal noise disrupts these features, complementing the aggregate raw-signal robustness results of Sec.~\ref{sec:robustness}.

\section{Results}\label{sec:results}

\begin{table*}[ht!]
    \centering
    \caption{Comparison of the small and large \TopTimeNet{} configurations selected by the hyperparameter search of Sec.~\ref{sec:hyp-search}. All test metrics are mean $\pm$ std over $30$ independent training runs. Both configurations use bilinear fusion (Sec.~\ref{sec:alg-training}), which has no rank or attention hyperparameters; the rank/attention-layer/attention-head values sampled elsewhere in the search do not apply to either selected configuration.}
    \label{tab:model-comparison}
    \setlength{\tabcolsep}{10pt}
    \begin{tabular}{lcc}
        \hline
        & \textbf{Small model} & \textbf{Large model} \\
        \hline
        Trainable parameters   & $1{,}638$            & $54{,}886$ \\
        Embedding dim.\ $D$    & $16$                 & $128$ \\
        Fusion strategy        & Bilinear             & Bilinear \\
        Classifier head        & $()$                 & $(64, 32, 16)$ \\
        Activation              & GELU                 & Leaky ReLU \\
        Dropout                 & $0.05$               & $0.0$ \\
        Optimizer              & AdamW                 & Adam \\
        Learning rate          & $7.12\times10^{-4}$  & $1.52\times10^{-6}$ \\
        Label smoothing        & $0.1$                & $0.0$ \\
        Gradient clip norm      & $5.0$                & $5.0$ \\
        Input normalization    & per-channel          & none \\
        \hline
        Accuracy               & $0.9758 \pm 0.0029$  & $0.9758 \pm 0.0022$ \\
        F1 score               & $0.9758 \pm 0.0029$  & $0.9757 \pm 0.0022$ \\
        G-mean                 & $0.9757 \pm 0.0029$  & $0.9756 \pm 0.0022$ \\
        Precision              & $0.9758 \pm 0.0029$  & $0.9757 \pm 0.0022$ \\
        Recall                 & $0.9758 \pm 0.0029$  & $0.9759 \pm 0.0022$ \\
        \hline
        Mean epochs trained    & $305.5 \pm 93.8$     & $500.0$ \\
        Mean training time (s) & $433.2 \pm 132.5$    & $873.9 \pm 11.5$ \\
        \hline
    \end{tabular}
\end{table*}

We evaluate \TopTimeNet{} on the full extended-teaspoon benchmark described in Sec.~\ref{sec:data_generation}, spanning $49$ nonlinear dynamical systems simulated in both periodic and chaotic regimes.

Before comparing against learned-representation baselines, we first ask a more basic question: given that the 
geometric and topological feature-extraction stage of \TopTimeNet{} is deterministic and contains no learned weights
(Sec.~\ref{sec:alg-features}), how much learnable capacity does the remaining discriminative stage actually need? Table~\ref{tab:model-comparison} compares two \TopTimeNet{} configurations selected by the hyperparameter search of Sec.~\ref{sec:hyp-search}: the globally best-scoring (``large'') configuration, with $54{,}886$ trainable parameters, and the parameter-efficient (``small'') configuration identified via the $1\%$-tolerance Pareto search described there, with $1{,}638$ parameters, approximately $33\times$ fewer. Despite this difference in trainable capacity, the two models achieve nearly identical mean test accuracy ($97.58\% \pm 0.29\%$ vs.\ $97.58\% \pm 0.22\%$, over $30$ independent training runs each).
The two configurations also yield very similar F1, G-mean, precision, and recall values.
The large model trained for the full $500$-epoch budget in every one of its $30$ runs, never triggering early stopping, whereas the small model stopped early at an average of epoch $306$ (range $140$--$500$).
Thus, under the adopted early-stopping criterion, training terminated earlier on average for the small configuration.
The small configuration therefore attains essentially the same mean predictive performance as the large configuration while using substantially fewer trainable parameters and terminating training earlier on average.

Under the present evaluation protocol, increasing the trainable parameter count from $1{,}638$ to $54{,}886$ does not improve the mean test metrics. This result is consistent with the fixed geometric and topological features already providing a representation from which the periodic and chaotic classes can be discriminated effectively with a comparatively compact learnable stage. The comparison does not, however, establish linear separability of the feature representation or a general capacity threshold beyond which additional parameters cannot be beneficial.
Guided by this result, we adopt the smaller, $1{,}638$-parameter configuration for all subsequent analyses in this section, including the comparison against convolutional and transformer-based baselines in Sec.~\ref{sec:baselines} and the robustness evaluation of Sec.~\ref{sec:robustness}.

\subsection{Comparison against convolutional and transformer-based baselines}\label{sec:baselines}

\begin{table*}[ht!]
    \centering
    \caption{\TopTimeNet{} (small configuration) versus CNN and Transformer baselines. \TopTimeNet{} and CNN report mean $\pm$ std over $30$ and $10$ independent training runs, respectively, with no failed or collapsed runs in either case. 
    For the Transformer, $10$ independent training attempts were performed; the reported mean $\pm$ std values are computed over the $7$ runs that converged, while the remaining $3$ attempts remained at a near-chance validation-accuracy plateau and are reported as non-converged (see text).}
    \label{tab:baselines}
    \setlength{\tabcolsep}{8pt}
    \begin{tabular}{lccc}
        \hline
        & \textbf{\TopTimeNet{} (small)} & \textbf{CNN ($n=10$)} & \textbf{Transformer ($n=7$)} \\
        \hline
        Trainable parameters & $1{,}638$           & $1{,}824{,}898$        & $33{,}435{,}570$ \\
        Accuracy             & $0.9758 \pm 0.0029$ & $0.9708 \pm 0.0110$    & $0.9412 \pm 0.0133$ \\
        F1 score             & $0.9758 \pm 0.0029$ & $0.9708 \pm 0.0110$    & $0.9412 \pm 0.0133$ \\
        G-mean               & $0.9757 \pm 0.0029$ & $0.9708 \pm 0.0111$    & $0.9411 \pm 0.0133$ \\
        Precision            & $0.9758 \pm 0.0029$ & $0.9708 \pm 0.0110$    & $0.9412 \pm 0.0133$ \\
        Recall               & $0.9758 \pm 0.0029$ & $0.9710 \pm 0.0109$    & $0.9416 \pm 0.0132$ \\
        Epochs trained       & $305.5 \pm 93.8$    & $103.4 \pm 23.1$       & $144.3 \pm 69.2$ \\
        \hline
    \end{tabular}
\end{table*}

To benchmark \TopTimeNet{}'s performance, we compare it against two learned-representation baselines trained directly on the raw segmented time series: a 1D convolutional neural network (CNN) and a Transformer encoder. Both baselines were selected using the same random-search strategy, trial-validity rules, and Pareto-based final-selection procedure described in Sec.~\ref{sec:hyp-search}, but over their own architecture-appropriate search spaces rather than \TopTimeNet{}'s, and with $250$ sampled trials and an objective weight of $\alpha=0.7$ each, versus $400$ trials and $\alpha=0.6$ for \TopTimeNet{} itself; early-stopping patience was fixed at $50$ epochs for all three architectures, in both the search and the final evaluation reported here. Both baselines are evaluated over $10$ independent training runs each, mirroring the repeated-run protocol used for \TopTimeNet{} in Table~\ref{tab:model-comparison} (at reduced scale, given the substantially higher per-run training cost of both baselines).

Table~\ref{tab:baselines} summarizes the CNN and Transformer configurations found by the search, alongside the small \TopTimeNet{} configuration from Table~\ref{tab:model-comparison}. The CNN baseline uses a channel-multiplier architecture (base channels $64$, multipliers $[1,2,4,8]$, max pooling) trained with Muon~\cite{jordan2024muon}; the Transformer baseline uses a $5$-layer encoder (embedding dimension $128$, $8$ attention heads, feedforward dimension $256$) with a convolutional input embedding, trained with AdamW~\cite{loshchilov2017adamw}.

Across $10$ independent runs, the CNN baseline achieves a mean accuracy of $97.08\% \pm 1.10\%$, close to \TopTimeNet{}'s $97.58\% \pm 0.30\%$. 
All $10$ CNN runs achieve accuracies between $94.8\%$ and $98.0\%$, with no non-converged run observed.
The Transformer baseline, in contrast, 
shows greater run-to-run instability under the tested training protocol:
of its $10$ independent training attempts, $3$ never escaped a near-chance-accuracy plateau (validation accuracy fluctuating around $50\%$ to $51\%$ from early in training onward) and were 
terminated
by early stopping once validation performance failed to improve for the configured patience window, well before the $500$-epoch budget was reached.

We classify these $3$ attempts as non-converged based on their validation trajectories and report the performance statistics in Table~\ref{tab:baselines} over the remaining $7$ runs. The three non-converged attempts reached final validation accuracies of $51.4\%$, $51.3\%$, and $54.2\%$, each with F1 and G-mean scores substantially below what their accuracy alone would suggest, consistent with strongly imbalanced class predictions. The remaining, converged runs achieve a mean accuracy of $94.12\% \pm 1.33\%$. Thus, $3$ of the $10$ Transformer training attempts did not converge under the tested protocol, whereas no non-converged runs were observed among the $30$ \TopTimeNet{} runs or the $10$ CNN runs.

The Transformer's converged runs stop, on average, after fewer epochs than \TopTimeNet{} ($144.3 \pm 69.2$ vs.\ $305.5 \pm 93.8$), but after more epochs than the CNN ($144.3 \pm 69.2$ vs.\ $103.4 \pm 23.1$).

Precision, recall, F1, and accuracy are numerically similar for the CNN and \TopTimeNet{}, and the converged Transformer runs show the same qualitative pattern, with recall ($94.16\% \pm 1.32\%$) close to the corresponding accuracy and precision.

The parameter counts also show a substantial difference in trainable model size.
The CNN baseline requires $1{,}824{,}898$ trainable parameters, over $1{,}100\times$ more than \TopTimeNet{}'s $1{,}638$, while achieving a similar mean accuracy; no non-converged runs were observed among its $10$ training attempts.

The Transformer baseline 
requires $33{,}435{,}570$ parameters (over $20{,}000\times$ \TopTimeNet{}'s parameter count and roughly $18\times$ the CNN's) while achieving a lower mean accuracy than either alternative over its $7$ converged runs, 
with $3$ of its $10$ training attempts classified as non-converged.

This is the central efficiency argument of this paper in concrete terms: under the tested protocol, the small \TopTimeNet{} configuration achieves mean accuracy comparable to the CNN and higher than the mean of the converged Transformer runs while using three to four orders of magnitude fewer trainable parameters. For the Transformer specifically, $3$ of the $10$ training attempts did not converge. These results are consistent with the small-versus-large \TopTimeNet{} comparison in Sec.~\ref{sec:results}, in which increasing the trainable parameter count did not improve the mean test metrics.

\subsection{Robustness evaluation}\label{sec:robustness}
\begin{figure*}[ht!]
    \centering
    \includegraphics[width=\linewidth]{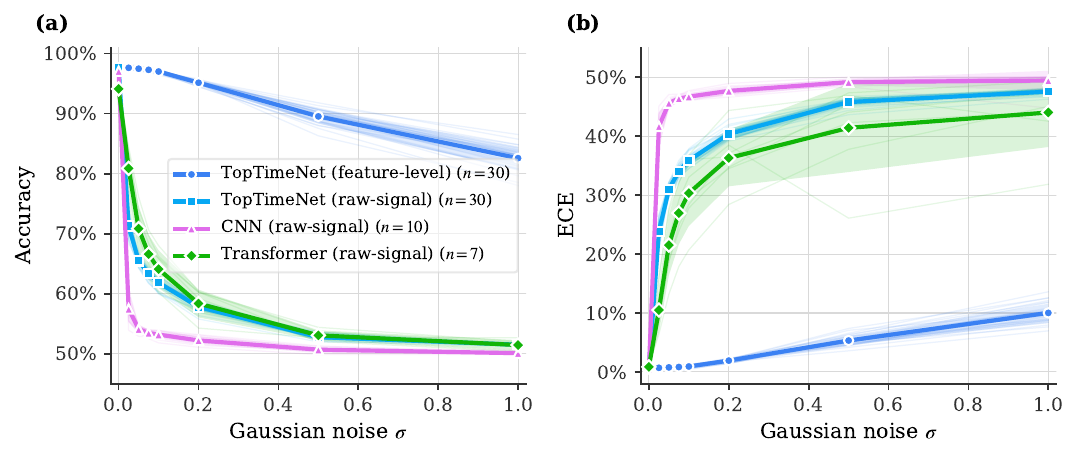}
    \caption{\textbf{Noise robustness of all three models.} \textbf{(a)} Test accuracy and \textbf{(b)} Expected Calibration Error (ECE, $M=10$ equal-width bins), both as a function of the Gaussian noise standard deviation $\sigma$. For \TopTimeNet{}, noise is injected at two distinct points: directly into the precomputed $42$-dimensional feature vector of Sec.~\ref{sec:alg-features} (``feature-level''), and into the raw time series itself, with the entire feature-extraction pipeline, Takens embedding, persistent homology, and all five feature branches, recomputed from the corrupted signal (``raw-signal''). The CNN and Transformer baselines have no intermediate feature representation to perturb separately, so only their raw-signal curves are shown. Thick lines show the mean over independent training runs ($n=30$ for \TopTimeNet{}, $n=10$ for the CNN, $n=7$ for the Transformer's converged runs), shaded bands show $\pm 1$ standard deviation, and thin lines show individual runs.}
    \label{fig:robustness}
\end{figure*}

Having established that the small, $1{,}638$-parameter \TopTimeNet{} configuration achieves mean clean-data accuracy comparable to the CNN and higher than the mean of the converged Transformer runs,
we now evaluate all three models' robustness to Gaussian noise. For \TopTimeNet{}, we distinguish two points at which noise can be introduced. In the \emph{feature-level} sweep, for each of the $30$ trained instances of the small model, we inject zero-mean Gaussian noise of standard deviation $\sigma$ directly into the held-out test set's precomputed $42$-dimensional feature vectors, reapply the same interquartile-range clipping bounds fit during feature precomputation, and evaluate the trained classifier on the resulting corrupted features. Because features are cached and reused across training runs, this sweep probes only the robustness of the learnable stage to 
perturbations of the precomputed geometric and topological summary features;
it does not exercise the Takens embedding or persistent homology computation under noise, and so cannot by itself establish the noise-sensitivity of the complete raw-signal-to-prediction pipeline.

In the \emph{raw-signal} sweep, we instead add Gaussian noise of the same standard deviation $\sigma$ directly to the held-out test segments themselves, and recompute the full $42$-dimensional feature vector from the noisy signal, including a fresh Vietoris--Rips persistent homology computation, before evaluating the same trained classifier, 
with the training-set-calibrated persistence-image grid held fixed across all noise levels.
The CNN and Transformer baselines have no analogous feature-level regime, since they consume the raw time series directly with no 
ana  ogous precomputed feature representationYou 
to perturb separately; noise injected into either baseline is therefore comparable in injection point to \TopTimeNet{}'s raw-signal sweep.

All sweeps use the same numerical noise grid ($\sigma \in \{0, 0.025, 0.05, 0.075, 0.1, 0.2, 0.5, 1.0\}$), applied as an absolute Gaussian noise standard deviation added directly to the raw signal or feature vector, rather than a signal-normalized noise level. Because the raw time series retain their native, system-specific amplitudes (Sec.~\ref{sec:training-protocol}), a given numerical $\sigma$ can correspond to substantially different noise levels relative to each system's or variable's own scale, as illustrated concretely for one example in Appendix~\ref{app:raw-signal-illustration}.
For \TopTimeNet{}, however, feature-level and raw-signal noise act in different spaces, so equal numerical values of $\sigma$ should not be interpreted as equal normalized perturbation strengths in the two sweeps. As illustrated concretely in Appendix~\ref{app:raw-signal-illustration}, the same nominal $\sigma$ can correspond to substantially different relative noise levels even between the two regimes of a single example: at $\sigma=0.1$, the injected noise there amounts to $25\%$ and $15\%$ of the periodic and chaotic signal's own standard deviation, respectively, while at $\sigma=0.5$ and $\sigma=1.0$ it reaches $73\%$--$127\%$ and $147\%$--$255\%$. The higher end of this noise grid is therefore not a subtle perturbation 
in either regime of this worked example: at $\sigma \geq 0.5$, the injected noise is comparable to or exceeds the clean signal's own scale, 
so these high-$\sigma$ points should not be interpreted as a uniform test of small-noise robustness across the benchmark.

The smallest tested noise levels are therefore more informative for assessing sensitivity to modest absolute perturbations, although their magnitude relative to the underlying signal still varies across systems and variables. 
The same independent training runs already reported in Table~\ref{tab:model-comparison} and Table~\ref{tab:baselines} are used throughout ($30$ for \TopTimeNet{}, $10$ for the CNN, and $7$ converged runs for the Transformer), 
so comparisons across noise levels do not involve different sets of trained instances within each model.

Only \TopTimeNet{} receives training-time noise augmentation, applied at the feature level ($\sigma_{\rm aug}=0.05$, doubling the effective training-set size), so it exposes the model only to noise in the $42$-dimensional feature space during training, never to noise in the raw signal.

The CNN and Transformer baselines, by contrast, receive no training-time noise augmentation of any kind; their raw-signal robustness sweep therefore evaluates a model trained exclusively on clean data, whereas \TopTimeNet{}'s feature-level sweep evaluates a model that was specifically trained to tolerate the kind of perturbation being tested.

Predicted-class confidences for all three models are calibrated post-hoc via a temperature $T$ fit on each run's clean validation set (mean $T = 0.545 \pm 0.024$ across \TopTimeNet{}'s $30$ runs) and reused unchanged across all noise levels for that run, so the calibration mapping itself is not refit as the noise level changes.

Fig.~\ref{fig:robustness} reports the resulting accuracy (panel (a)) and ECE (panel (b)) curves. At $\sigma=0$, all curves recover each model's clean-data accuracy from Tables~\ref{tab:model-comparison} and~\ref{tab:baselines}. \TopTimeNet{}'s feature-level curve is the only one of the four that degrades gracefully: accuracy remains above $97\%$ through $\sigma=0.1$ ($97.02\% \pm 0.25\%$), above $95\%$ at $\sigma=0.2$, and only falls substantially at the highest tested noise levels, reaching $89.55\% \pm 1.18\%$ at $\sigma=0.5$ and $82.58\% \pm 1.94\%$ at $\sigma=1.0$. All three raw-signal curves, by contrast, degrade sharply within the first few tested noise levels. \TopTimeNet{}'s own raw-signal accuracy falls to $71.27\% \pm 1.47\%$ already at $\sigma=0.025$, a drop of over $26$ percentage points from its clean-data value,
and plateaus near chance level ($51.51\% \pm 0.12\%$) by $\sigma=1.0$. The CNN 
shows the largest initial drop among the three raw-signal curves:
clean accuracy of $97.08\% \pm 1.10\%$ falls to $57.48\% \pm 2.23\%$ at $\sigma=0.025$ and reaches $50.10\% \pm 0.22\%$ by $\sigma=1.0$. The Transformer's converged runs show the same qualitative collapse but from a lower starting point and a comparatively smaller initial drop ($94.12\% \pm 1.33\%$ to $80.88\% \pm 3.38\%$ at $\sigma=0.025$), reaching a similar chance-level floor ($51.45\% \pm 0.82\%$) by $\sigma=1.0$. 
Thus, all three tested pipelines show substantial sensitivity to raw-signal Gaussian perturbations under the present protocol, despite their different architectures and parameter counts.

Panel (b) shows the corresponding contrast in calibration. \TopTimeNet{}'s feature-level ECE increases from $0.69\%$ at $\sigma=0$ to $10.02\% \pm 1.54\%$ at $\sigma=1.0$. All three raw-signal curves climb much more steeply: \TopTimeNet{}'s own raw-signal ECE jumps to $23.86\% \pm 1.71\%$ at $\sigma=0.025$ 
and reaches $47.54\% \pm 0.45\%$ by $\sigma=1.0$; the CNN reaches $41.65\% \pm 2.35\%$ already at $\sigma=0.025$ and $49.42\% \pm 1.65\%$ by $\sigma=1.0$; and the Transformer's converged runs reach $44.01\% \pm 5.86\%$ by $\sigma=1.0$. 
At the highest raw-signal noise levels, all three models therefore combine near-chance accuracy with large ECE values, indicating substantial degradation in the calibration of their predicted probabilities.

For \TopTimeNet{}, this contrast shows that robustness to perturbations of the precomputed feature vector does not imply robustness when noise is introduced in the raw signal and the features are recomputed.
Because the feature-level and raw-signal sweeps perturb different spaces, and because \TopTimeNet{} is trained with feature-level but not raw-signal noise augmentation, the present experiment does not isolate the source of the difference between these two robustness curves. Independent of any effect of training-time augmentation, Appendix~\ref{app:raw-signal-illustration} shows directly, via a single worked example, that the delay-embedded point cloud and its persistence diagram are themselves visibly disrupted by raw-signal noise at these same $\sigma$ levels, before any classifier is involved: raw-signal noise can disrupt the representation itself, independent of the classifier, though this one example does not indicate what fraction of the aggregate, benchmark-level degradation this mechanism accounts for.

For deployment, these results emphasize that robustness to perturbations of precomputed features should not be taken as evidence of robustness to noise present at acquisition time. Such noise should instead be evaluated by propagating the corrupted signal through the complete signal-to-prediction pipeline. See Appendix~\ref{app:raw-signal-illustration} for a single-example illustration of this mechanism.

\section{Conclusion}\label{sec:conclusion}

We introduced \TopTimeNet{}, a time-series classification architecture that separates fixed feature construction from discrimination: five families of features derived from Takens delay embeddings and persistent homology, geometric, entropy, lifetime, Betti curve, and persistence image statistics, are computed deterministically and without any trainable parameters, leaving only a lightweight learnable stage to solve the resulting classification task. Using the double pendulum as an illustrative example, we showed how the five branches provide complementary descriptions of the contrast between periodic and chaotic trajectories, and showed on the nonlinear-system benchmark described in Sec.~\ref{sec:data_generation} that a $1{,}638$-parameter classifier achieves the same mean accuracy as a configuration with $33\times$ more trainable parameters, and achieves mean accuracy comparable to the CNN and higher than the mean of the converged Transformer runs, while using three to four orders of magnitude fewer trainable parameters.

Reliability across repeated training runs varied by baseline: 
no non-converged runs were observed among the $30$ \TopTimeNet{} runs or the $10$ CNN runs, whereas $3$ of the $10$ Transformer training attempts did not converge under the tested protocol.

Our noise experiments distinguish robustness to perturbations of precomputed features from robustness of the complete raw-signal-to-prediction pipeline.
Perturbing the precomputed feature vectors directly, \TopTimeNet{} degrades gracefully, retaining most of its accuracy across a wide range of injected noise. Perturbing the raw time series instead, and recomputing the full delay-embedding-to-persistent-homology pipeline on the corrupted signal, produces a markedly different outcome: accuracy degrades sharply at the smallest tested absolute noise levels, although the magnitude of a given $\sigma$ relative to the underlying signal varies across systems and variables (Appendix~\ref{app:raw-signal-illustration}).

The CNN and Transformer baselines, which have no analogous precomputed feature representation to perturb separately, show the same qualitative collapse under raw-signal noise. Thus, all three tested pipelines are sensitive to raw-signal Gaussian perturbations under the present protocol, despite their different architectures and parameter counts.

For \TopTimeNet{}, robustness to perturbations of the precomputed feature vector therefore does not imply robustness when noise is introduced before feature extraction. Because the two experiments perturb different spaces, and because feature-level but not raw-signal noise is used for \TopTimeNet{}'s training augmentation, the present experiments do not isolate the source of the difference between the two robustness curves.

These observations do not conflict with standard stability results for Vietoris--Rips persistence, which control changes in persistence diagrams under suitable perturbations of the underlying metric space~\cite{chazal2014persistence}. Such results do not guarantee stability of the complete pipeline, including delay embedding, finite-sample summary statistics, preprocessing, and classification. In particular, the Gaussian perturbations considered here are added to the observed time series after trajectory generation and should not be interpreted as perturbations of the initial conditions whose effects are subsequently amplified by chaotic dynamics.
Robustness to noise present at the point of raw-signal acquisition therefore cannot be assumed to follow automatically from persistent homology's stability properties, and should instead be established empirically, as we have done here.

Appendix~\ref{app:raw-signal-illustration} illustrates one contributing mechanism concretely: for a single worked example, the periodic regime's topological signature, a single dominant, long-lived loop, is more easily destroyed by observational noise than the chaotic regime's already-diffuse signature, consistent with both the periodic signal's smaller natural amplitude (making a given noise level proportionally larger) and its simpler, lower-entropy structure. We emphasize that this single-example asymmetry illustrates a candidate mechanism rather than a general claim that periodic dynamics are always more fragile than chaotic dynamics to observational noise.

The present study is deliberately scoped to a binary distinction between periodic and chaotic dynamics. Several directions follow naturally from this scope. Most immediately, the same feature set could plausibly extend to distinguishing a broader range of dynamical regimes beyond the periodic/chaotic dichotomy considered here. 
Quasi-periodic and stochastic dynamics provide natural additional test cases because their reconstructed trajectories may exhibit geometric and topological structure different from the periodic and chaotic examples considered here. Whether the present fixed embedding and feature set can separate these regimes reliably remains to be established. Testing \TopTimeNet{} on a labeled multi-class benchmark spanning periodic, quasi-periodic, chaotic, and stochastic regimes would be a natural way to evaluate these hypotheses.

A second direction concerns quantum dynamics. 
Topological data analysis has been applied to quantum dynamics in several contexts, including regular--chaotic discrimination and persistent-homology-based monitoring of finite-time quantum engines~\cite{Cao2023, KeremMaden_2026},
and we hypothesize that the same TDA-based feature branches used here for classical dynamical systems could be similarly informative for classifying periodic versus chaotic quantum dynamics, for instance from time series of expectation values, wavefunction overlaps, or other quantum observables whose reconstructed phase-space embeddings might exhibit analogous topological signatures of regularity and chaos. We view this as a promising
direction for testing the present framework, rather than an outcome we take for granted.

More broadly, the underlying idea of this work, replacing a portion of a model's learned representation with a fixed, domain-informed feature extractor and allowing a small number of trainable parameters to focus solely on the discriminative task, is not specific to periodic/chaotic classification, or even to time-series data. 
The same representation/discrimination decoupling could be tested in other learning tasks for which informative domain-based feature representations are available.
Testing this hypothesis in other domains, and characterizing where the resulting efficiency benefits and robustness behavior do and do not carry over,
remains an open and, we believe, worthwhile direction for future work.

\section{Code availability statement}

The code that supports the findings of this article, along with the
scripts and instructions needed to regenerate the dataset used in our
experiments, is openly available in the
\href{https://github.com/say-yas/TopTimeNet.git}{\TopTimeNet{} GitHub repository}.

\section{Author contributions}
S.~S. conceived the project, implemented the algorithm, analyzed the data and prepared the initial draft of the manuscript. S.~B. contributed to the analysis of the results and together with S.~S. finalized the manuscript. 

\section{Acknowledgment}
S.~S. thanks B.~Krishnamoorthy for helpful communications at early statges of this project. We acknowledge the use of the following open-source packages in the preparation of our code and analysis: Teaspoon~\cite{Khasawneh2025}, Ripser.py~\cite{ctralie2018ripser}, Scikit-TDA~\cite{scikittda2019}, and GUDHI~\cite{gudhi:PersistenceRepresentations}. In preparing this manuscript, we benefited from
Anthropic’s Claude Sonnet and OpenAI's ChatGPT (GPT-5.6 Sol) for polishing and improving the readability of the text. The authors remain responsible for the content.

\appendix
\section{Illustrative raw-signal noise sweep on the double pendulum}\label{app:raw-signal-illustration}

Sec.~\ref{sec:robustness} quantifies the raw-signal noise sweep in aggregate, averaged over $30$ independently trained classifiers across all $49$ systems in the benchmark. Here we illustrate the same sweep on a single example, the double pendulum, to give visual intuition for the underlying mechanism. Using the feature-extraction pipeline of Sec.~\ref{sec:example}, we add zero-mean Gaussian noise of standard deviation $\sigma$ directly to one periodic and one chaotic double-pendulum segment, and recompute the full pipeline, Takens embedding, persistent homology, and all five feature branches, from the corrupted signal at each of five representative noise levels, $\sigma \in \{0, 0.1, 0.2, 0.5, 1.0\}$.

\begin{figure*}[ht!]
    \centering
    \includegraphics[width=\linewidth]{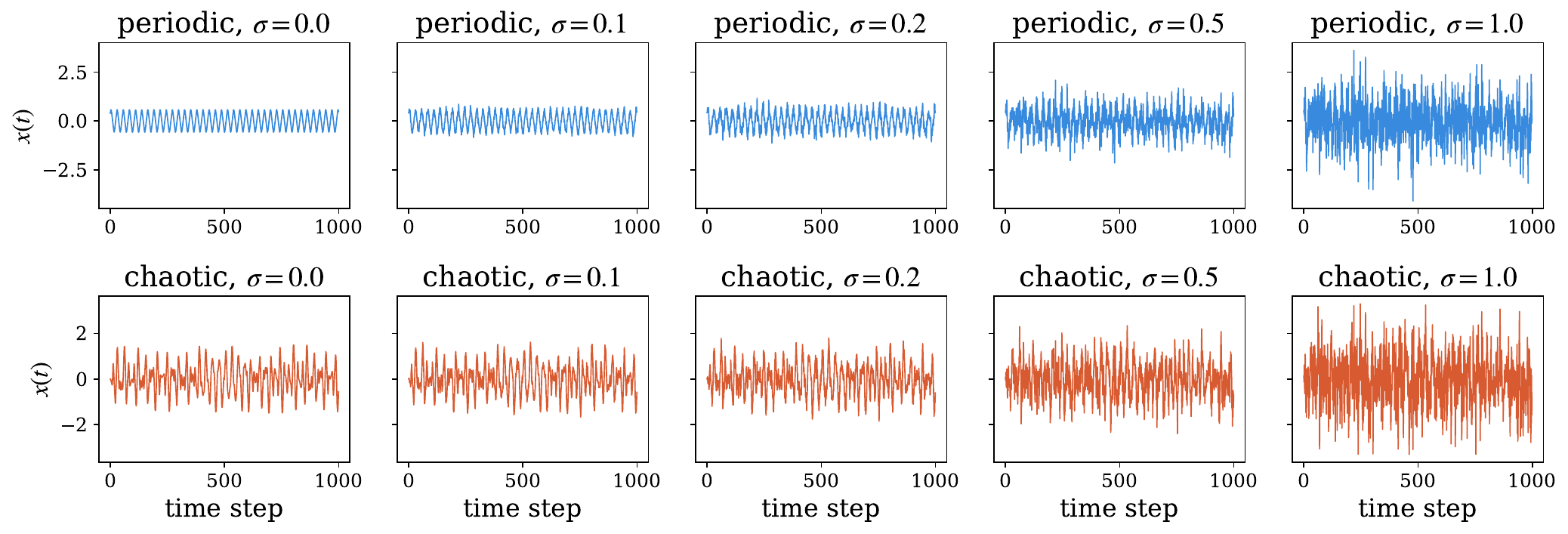}
    \caption{\textbf{Raw double-pendulum time series under increasing Gaussian noise.} Top: periodic. Bottom: chaotic.}
    \label{fig:app-timeseries}
\end{figure*}

Because $\sigma$ is an absolute noise standard deviation while the paper's SNR convention (Sec.~\ref{sec:robustness}) is defined relative to a unit reference power, it is worth stating what these nominal $\sigma$ values represent relative to this specific example's own signal amplitude. We define the \emph{relative noise level} of a given $\sigma$, for a clean signal $x(t)$, as
$ r(\sigma) = \sigma / \mathrm{std}(x), $
the ratio of the injected noise's own standard deviation to the clean signal's standard deviation; $r(\sigma)=1$ (i.e., $100\%$) corresponds to noise exactly as large, in this sense, as the signal itself. The clean periodic segment has $\mathrm{std}(x)=0.39$ and the clean chaotic segment has $\mathrm{std}(x)=0.68$; across the four noise levels shown, $r(\sigma)$ is $25\%$/$15\%$ at $\sigma=0.1$, $51\%$/$29\%$ at $\sigma=0.2$, $127\%$/$73\%$ at $\sigma=0.5$, and $255\%$/$147\%$ at $\sigma=1.0$ (periodic/chaotic, respectively). The injected noise therefore stays below the signal's own scale for both regimes only at $\sigma=0.1$; by $\sigma=0.5$ it already exceeds the periodic signal's own scale and approaches the chaotic signal's, and at $\sigma=1.0$ it exceeds both, but considerably more so for periodic than for chaotic throughout.

Fig.~\ref{fig:app-timeseries} shows the corrupted raw signals themselves. At $\sigma=0.1$ ($r(\sigma) \le 25\%$ for both regimes), the underlying oscillatory structure remains clearly visible to the eye in both signals, with only a modest increase in high-frequency texture; at $\sigma=0.2$ this texture is more pronounced but the oscillation is still readily apparent. Only at $\sigma=0.5$ and above, where the injected noise is comparable to or exceeds the signal's own scale, does the raw signal begin to look obviously noisy, with the periodic signal's regular oscillation becoming visually difficult to discern by $\sigma=1.0$. This is precisely the point: the downstream topological representation degrades well before the raw signal appears corrupted, and well before the noise is large relative to the signal itself.

\begin{figure*}[ht!]
    \centering
    \includegraphics[width=\linewidth]{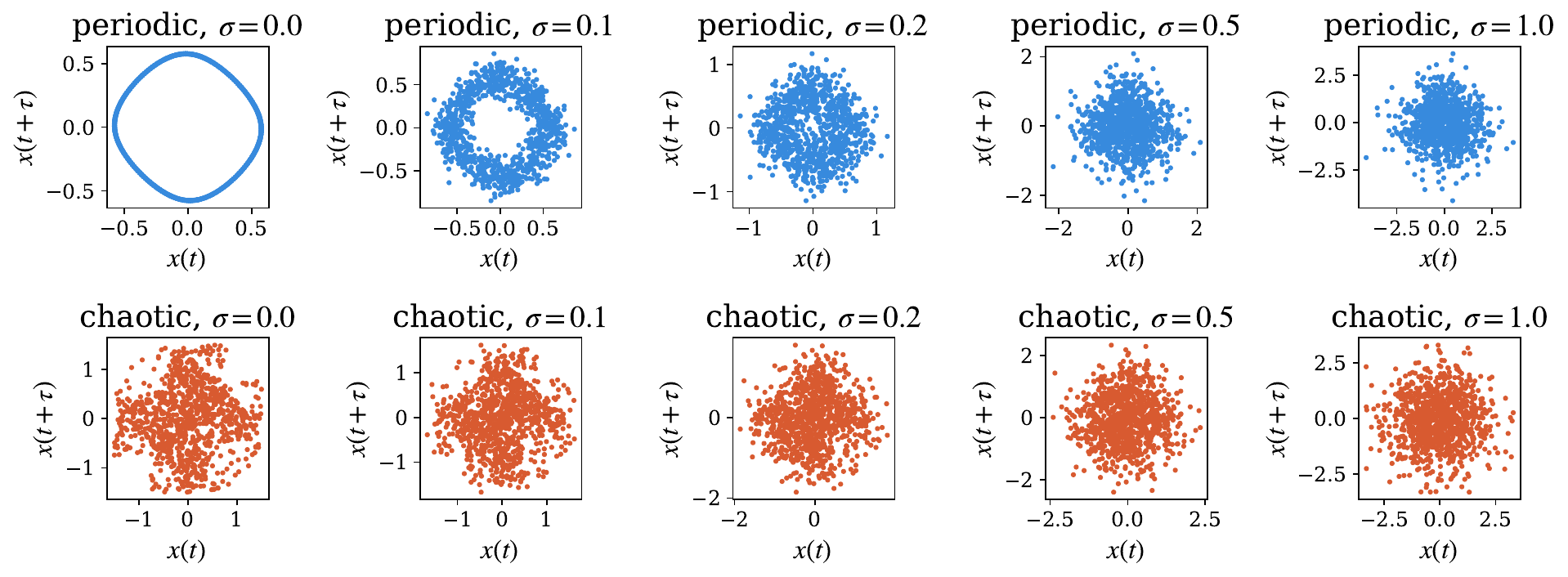}
    \caption{\textbf{Takens-embedded point clouds under raw-signal Gaussian noise.} Top: periodic. Bottom: chaotic.}
    \label{fig:app-pointclouds}
\end{figure*}

Fig.~\ref{fig:app-pointclouds} shows the resulting point clouds. At $\sigma=0$, the periodic trajectory embeds as a clean, thin ring, a simple limit cycle. At $\sigma=0.1$, this ring has visibly thickened but a central hole, the signature of a genuine loop, remains discernible; by $\sigma=0.2$ that hole has effectively closed, and the point cloud has already collapsed into a diffuse blob with no remaining loop structure, a state essentially unchanged through $\sigma=0.5$ and $\sigma=1.0$. The chaotic trajectory's point cloud, by contrast, is already a diffuse, space-filling scatter at $\sigma=0$, so it has comparatively less clean structure to lose, and its qualitative appearance changes far less over the same noise range. Part of this asymmetry is attributable to the differing relative noise levels noted above (the same nominal $\sigma$ is a proportionally larger perturbation for the periodic signal, given its smaller natural amplitude), though the periodic point cloud's simpler, lower-dimensional structure, a thin ring rather than a space-filling scatter, plausibly also makes it more visually sensitive to a comparable relative perturbation.

\begin{figure*}[ht!]
    \centering
    \includegraphics[width=\linewidth]{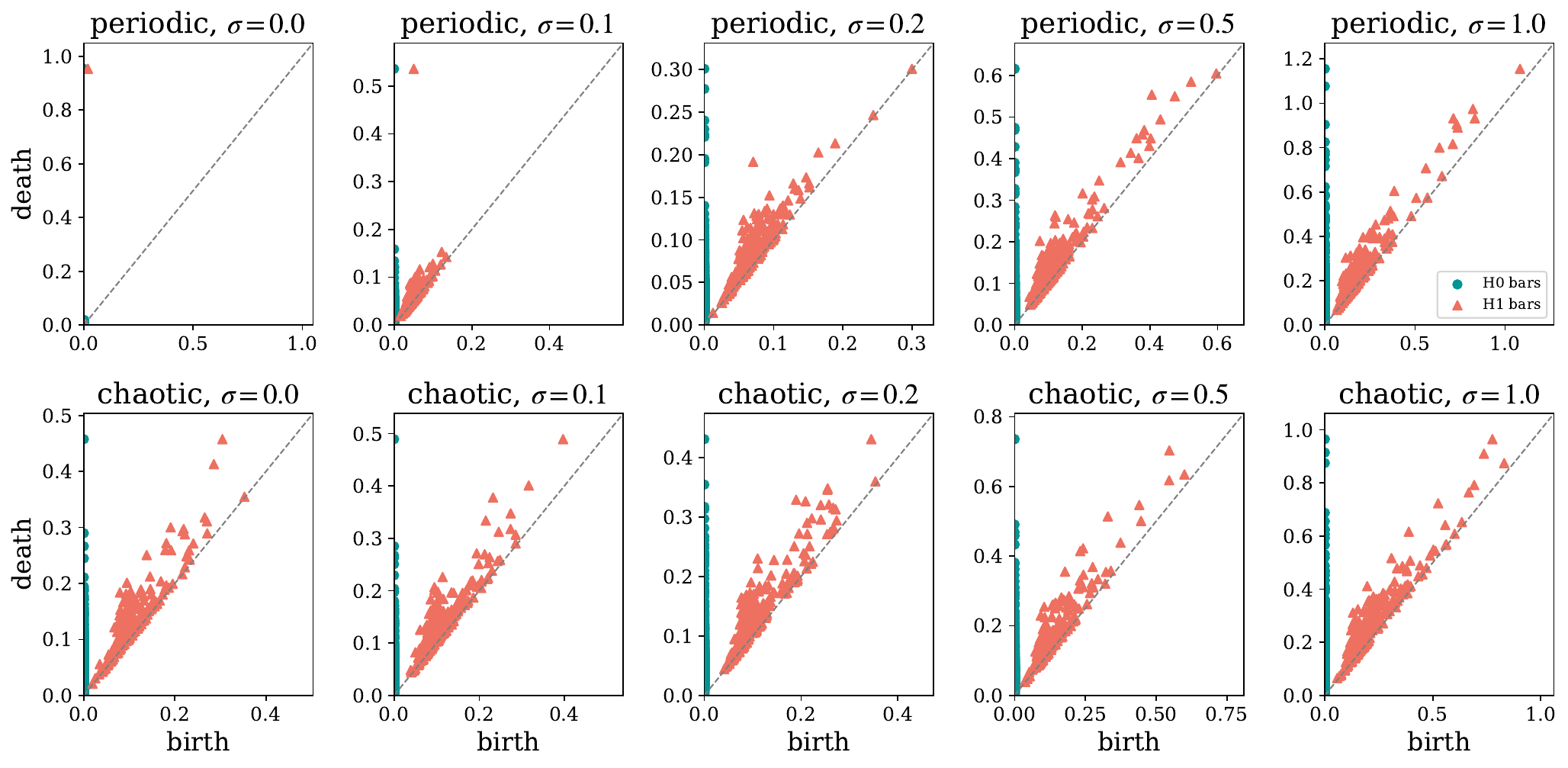}
    \caption{\textbf{Persistence diagrams under raw-signal Gaussian noise.} Top: periodic. Bottom: chaotic.}
    \label{fig:app-diagrams}
\end{figure*}

Fig.~\ref{fig:app-diagrams} shows the corresponding persistence diagrams. At $\sigma=0$, the periodic regime's diagram contains a single dominant, long-lived $H_1$ bar and essentially nothing else, exactly the clean-loop signature visible in Fig.~\ref{fig:app-pointclouds}. As $\sigma$ increases, this dominant bar's persistence steadily shrinks, from a death time near $0.95$ at $\sigma=0$ to roughly $0.5$ at $\sigma=0.1$ and roughly $0.3$ at $\sigma=0.2$, while a growing cloud of small, noise-induced $H_1$ bars appears near the diagonal; by $\sigma=0.5$ this cloud has grown large enough that the once-dominant bar is only marginally distinguishable from it. The chaotic regime's diagram already contains many scattered $H_1$ bars of comparable persistence at $\sigma=0$, and this qualitative pattern is largely preserved across the full noise range, aside from an overall rescaling of the birth/death axes.

\begin{figure}[ht!]
    \centering
    \includegraphics[width=0.8\linewidth]{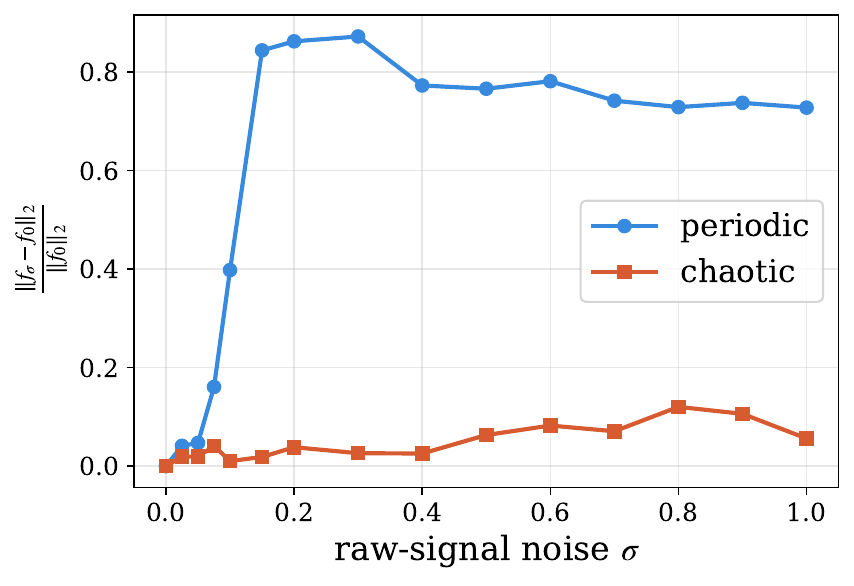}
    \caption{\textbf{Relative feature-vector drift} $\|f_\sigma - f_0\|_2 / \|f_0\|_2$ \textbf{under raw-signal Gaussian noise}, for the periodic and chaotic double-pendulum examples of Figs.~\ref{fig:app-timeseries}--\ref{fig:app-diagrams}.}
    \label{fig:app-drift}
\end{figure}

Fig.~\ref{fig:app-drift} shows the corresponding relative feature-vector drift, $\|f_\sigma - f_0\|_2 / \|f_0\|_2$, where $f_\sigma \in \mathbb{R}^{42}$ denotes the full feature vector of Sec.~\ref{sec:alg-features} recomputed from the signal at noise level $\sigma$ (and $f_0$ the same quantity computed on the clean signal). The periodic curve rises sharply between $\sigma=0.05$ and $\sigma=0.15$, from below $0.1$ to above $0.8$, then plateaus around $0.7$--$0.85$ for all larger $\sigma$ tested; the chaotic curve remains below $0.13$ across the entire range. The periodic example's feature vector thus moves substantially further from its clean value than the chaotic example's does at every tested $\sigma$ (e.g., $0.40$ vs.\ $0.03$ at $\sigma=0.1$, $0.86$ vs.\ $0.04$ at $\sigma=0.2$, and $0.73$ vs.\ $0.06$ at $\sigma=1.0$), consistent with both the relative-noise-level asymmetry and the qualitative point-cloud and persistence-diagram observations above.

We emphasize that this single-example asymmetry, periodic proving more fragile than chaotic here, illustrates \emph{a} mechanism by which raw-signal noise degrades a distinguishing topological signature, not a claim that periodic dynamics are universally more noise-sensitive than chaotic dynamics; both the specific dynamical system's signal amplitude and its topological structure plausibly contribute, and either could dominate for a different system or parameter regime. The paper's aggregate finding, that raw-signal noise collapses classification accuracy far more sharply than feature-level noise of the same magnitude, is established in Sec.~\ref{sec:robustness} by averaging over $30$ independently trained classifiers and is the result that should be treated as representative; this appendix is included only to make the underlying mechanism visually concrete.

 \bibliography{bibfile}

\begin{thebibliography}{46}%
\makeatletter
\providecommand \@ifxundefined [1]{%
 \@ifx{#1\undefined}
}%
\providecommand \@ifnum [1]{%
 \ifnum #1\expandafter \@firstoftwo
 \else \expandafter \@secondoftwo
 \fi
}%
\providecommand \@ifx [1]{%
 \ifx #1\expandafter \@firstoftwo
 \else \expandafter \@secondoftwo
 \fi
}%
\providecommand \natexlab [1]{#1}%
\providecommand \enquote  [1]{``#1''}%
\providecommand \bibnamefont  [1]{#1}%
\providecommand \bibfnamefont [1]{#1}%
\providecommand \citenamefont [1]{#1}%
\providecommand \href@noop [0]{\@secondoftwo}%
\providecommand \href [0]{\begingroup \@sanitize@url \@href}%
\providecommand \@href[1]{\@@startlink{#1}\@@href}%
\providecommand \@@href[1]{\endgroup#1\@@endlink}%
\providecommand \@sanitize@url [0]{\catcode `\\12\catcode `\$12\catcode
  `\&12\catcode `\#12\catcode `\^12\catcode `\_12\catcode `\%12\relax}%
\providecommand \@@startlink[1]{}%
\providecommand \@@endlink[0]{}%
\providecommand \url  [0]{\begingroup\@sanitize@url \@url }%
\providecommand \@url [1]{\endgroup\@href {#1}{\urlprefix }}%
\providecommand \urlprefix  [0]{URL }%
\providecommand \Eprint [0]{\href }%
\providecommand \doibase [0]{https://doi.org/}%
\providecommand \selectlanguage [0]{\@gobble}%
\providecommand \bibinfo  [0]{\@secondoftwo}%
\providecommand \bibfield  [0]{\@secondoftwo}%
\providecommand \translation [1]{[#1]}%
\providecommand \BibitemOpen [0]{}%
\providecommand \bibitemStop [0]{}%
\providecommand \bibitemNoStop [0]{.\EOS\space}%
\providecommand \EOS [0]{\spacefactor3000\relax}%
\providecommand \BibitemShut  [1]{\csname bibitem#1\endcsname}%
\let\auto@bib@innerbib\@empty
\bibitem [{\citenamefont {Provenzale}\ \emph {et~al.}(1992)\citenamefont
  {Provenzale}, \citenamefont {Smith}, \citenamefont {Vio},\ and\ \citenamefont
  {Murante}}]{Provenzale1992}%
  \BibitemOpen
  \bibfield  {author} {\bibinfo {author} {\bibfnamefont {A.}~\bibnamefont
  {Provenzale}}, \bibinfo {author} {\bibfnamefont {L.~A.}\ \bibnamefont
  {Smith}}, \bibinfo {author} {\bibfnamefont {R.}~\bibnamefont {Vio}},\ and\
  \bibinfo {author} {\bibfnamefont {G.}~\bibnamefont {Murante}},\ }\bibfield
  {title} {\bibinfo {title} {Distinguishing between low-dimensional dynamics
  and randomness in measured time series},\ }\href
  {https://doi.org/10.1016/0167-2789(92)90100-2} {\bibfield  {journal}
  {\bibinfo  {journal} {Physica D: Nonlinear Phenomena}\ }\textbf {\bibinfo
  {volume} {58}},\ \bibinfo {pages} {31} (\bibinfo {year} {1992})}\BibitemShut
  {NoStop}%
\bibitem [{\citenamefont {Tsonis}\ and\ \citenamefont
  {Elsner}(1992)}]{Tsonis1992}%
  \BibitemOpen
  \bibfield  {author} {\bibinfo {author} {\bibfnamefont {A.~A.}\ \bibnamefont
  {Tsonis}}\ and\ \bibinfo {author} {\bibfnamefont {J.~B.}\ \bibnamefont
  {Elsner}},\ }\bibfield  {title} {\bibinfo {title} {Nonlinear prediction as a
  way of distinguishing chaos from random fractal sequences},\ }\href
  {https://doi.org/10.1038/358217a0} {\bibfield  {journal} {\bibinfo  {journal}
  {Nature}\ }\textbf {\bibinfo {volume} {358}},\ \bibinfo {pages} {217}
  (\bibinfo {year} {1992})}\BibitemShut {NoStop}%
\bibitem [{\citenamefont {Cencini}\ \emph {et~al.}(2000)\citenamefont
  {Cencini}, \citenamefont {Falcioni}, \citenamefont {Olbrich}, \citenamefont
  {Kantz},\ and\ \citenamefont {Vulpiani}}]{Cencini2000}%
  \BibitemOpen
  \bibfield  {author} {\bibinfo {author} {\bibfnamefont {M.}~\bibnamefont
  {Cencini}}, \bibinfo {author} {\bibfnamefont {M.}~\bibnamefont {Falcioni}},
  \bibinfo {author} {\bibfnamefont {E.}~\bibnamefont {Olbrich}}, \bibinfo
  {author} {\bibfnamefont {H.}~\bibnamefont {Kantz}},\ and\ \bibinfo {author}
  {\bibfnamefont {A.}~\bibnamefont {Vulpiani}},\ }\bibfield  {title} {\bibinfo
  {title} {Chaos or noise: Difficulties of a distinction},\ }\href
  {https://doi.org/10.1103/PhysRevE.62.427} {\bibfield  {journal} {\bibinfo
  {journal} {Physical Review E}\ }\textbf {\bibinfo {volume} {62}},\ \bibinfo
  {pages} {427} (\bibinfo {year} {2000})}\BibitemShut {NoStop}%
\bibitem [{\citenamefont {Grassberger}\ and\ \citenamefont
  {Procaccia}(1983{\natexlab{a}})}]{Grassberger1983b}%
  \BibitemOpen
  \bibfield  {author} {\bibinfo {author} {\bibfnamefont {P.}~\bibnamefont
  {Grassberger}}\ and\ \bibinfo {author} {\bibfnamefont {I.}~\bibnamefont
  {Procaccia}},\ }\bibfield  {title} {\bibinfo {title} {Characterization of
  strange attractors},\ }\href {https://doi.org/10.1103/PhysRevLett.50.346}
  {\bibfield  {journal} {\bibinfo  {journal} {Phys. Rev. Lett.}\ }\textbf
  {\bibinfo {volume} {50}},\ \bibinfo {pages} {346} (\bibinfo {year}
  {1983}{\natexlab{a}})}\BibitemShut {NoStop}%
\bibitem [{\citenamefont {Casaleggio}\ \emph {et~al.}(1995)\citenamefont
  {Casaleggio}, \citenamefont {Corana},\ and\ \citenamefont
  {Ridella}}]{casaleggio1995}%
  \BibitemOpen
  \bibfield  {author} {\bibinfo {author} {\bibfnamefont {A.}~\bibnamefont
  {Casaleggio}}, \bibinfo {author} {\bibfnamefont {A.}~\bibnamefont {Corana}},\
  and\ \bibinfo {author} {\bibfnamefont {S.}~\bibnamefont {Ridella}},\
  }\bibfield  {title} {\bibinfo {title} {Correlation dimension estimation from
  electrocardiograms},\ }\href
  {https://doi.org/https://doi.org/10.1016/0960-0779(93)E0053-E} {\bibfield
  {journal} {\bibinfo  {journal} {Chaos, Solitons \& Fractals}\ }\textbf
  {\bibinfo {volume} {5}},\ \bibinfo {pages} {713} (\bibinfo {year}
  {1995})}\BibitemShut {NoStop}%
\bibitem [{\citenamefont {Argyris}\ \emph {et~al.}(1998)\citenamefont
  {Argyris}, \citenamefont {Andreadis}, \citenamefont {Pavlos},\ and\
  \citenamefont {Athanasiou}}]{argyris1998}%
  \BibitemOpen
  \bibfield  {author} {\bibinfo {author} {\bibfnamefont {J.}~\bibnamefont
  {Argyris}}, \bibinfo {author} {\bibfnamefont {I.}~\bibnamefont {Andreadis}},
  \bibinfo {author} {\bibfnamefont {G.}~\bibnamefont {Pavlos}},\ and\ \bibinfo
  {author} {\bibfnamefont {M.}~\bibnamefont {Athanasiou}},\ }\bibfield  {title}
  {\bibinfo {title} {The influence of noise on the correlation dimension of
  chaotic attractors},\ }\href
  {https://doi.org/https://doi.org/10.1016/S0960-0779(97)00120-3} {\bibfield
  {journal} {\bibinfo  {journal} {Chaos, Solitons \& Fractals}\ }\textbf
  {\bibinfo {volume} {9}},\ \bibinfo {pages} {343} (\bibinfo {year}
  {1998})}\BibitemShut {NoStop}%
\bibitem [{\citenamefont {Boull{\'e}}\ \emph {et~al.}(2020)\citenamefont
  {Boull{\'e}}, \citenamefont {Dallas}, \citenamefont {Nakatsukasa},\ and\
  \citenamefont {Samaddar}}]{boulle2020classification}%
  \BibitemOpen
  \bibfield  {author} {\bibinfo {author} {\bibfnamefont {N.}~\bibnamefont
  {Boull{\'e}}}, \bibinfo {author} {\bibfnamefont {V.}~\bibnamefont {Dallas}},
  \bibinfo {author} {\bibfnamefont {Y.}~\bibnamefont {Nakatsukasa}},\ and\
  \bibinfo {author} {\bibfnamefont {D.}~\bibnamefont {Samaddar}},\ }\bibfield
  {title} {\bibinfo {title} {Classification of chaotic time series with deep
  learning},\ }\href
  {https://doi.org/https://doi.org/10.1016/j.physd.2019.132261} {\bibfield
  {journal} {\bibinfo  {journal} {Physica D: Nonlinear Phenomena}\ }\textbf
  {\bibinfo {volume} {403}},\ \bibinfo {pages} {132261} (\bibinfo {year}
  {2020})}\BibitemShut {NoStop}%
\bibitem [{\citenamefont {Zanin}(2022)}]{Zanin2022}%
  \BibitemOpen
  \bibfield  {author} {\bibinfo {author} {\bibfnamefont {M.}~\bibnamefont
  {Zanin}},\ }\bibfield  {title} {\bibinfo {title} {Can deep learning
  distinguish chaos from noise? numerical experiments and general
  considerations},\ }\href {https://doi.org/10.1016/j.cnsns.2022.106708}
  {\bibfield  {journal} {\bibinfo  {journal} {Communications in Nonlinear
  Science and Numerical Simulation}\ }\textbf {\bibinfo {volume} {114}},\
  \bibinfo {pages} {106708} (\bibinfo {year} {2022})}\BibitemShut {NoStop}%
\bibitem [{\citenamefont {Choi}\ \emph {et~al.}(2026)\citenamefont {Choi},
  \citenamefont {Chanu},\ and\ \citenamefont {Park}}]{choi2026learning}%
  \BibitemOpen
  \bibfield  {author} {\bibinfo {author} {\bibfnamefont {J.}~\bibnamefont
  {Choi}}, \bibinfo {author} {\bibfnamefont {A.~L.}\ \bibnamefont {Chanu}},\
  and\ \bibinfo {author} {\bibfnamefont {J.-M.}\ \bibnamefont {Park}},\
  }\bibfield  {title} {\bibinfo {title} {Learning a quantitative criterion for
  distinguishing chaos from noise},\ }\href@noop {} {\bibfield  {journal}
  {\bibinfo  {journal} {arXiv:2608.07109}\ } (\bibinfo {year}
  {2026})}\BibitemShut {NoStop}%
\bibitem [{\citenamefont {Wang}\ \emph {et~al.}(2017)\citenamefont {Wang},
  \citenamefont {Yan},\ and\ \citenamefont {Oates}}]{Wang2017}%
  \BibitemOpen
  \bibfield  {author} {\bibinfo {author} {\bibfnamefont {Z.}~\bibnamefont
  {Wang}}, \bibinfo {author} {\bibfnamefont {W.}~\bibnamefont {Yan}},\ and\
  \bibinfo {author} {\bibfnamefont {T.}~\bibnamefont {Oates}},\ }\bibfield
  {title} {\bibinfo {title} {Time series classification from scratch with deep
  neural networks: A strong baseline},\ }in\ \href
  {https://doi.org/10.1109/IJCNN.2017.7966039} {\emph {\bibinfo {booktitle}
  {2017 International Joint Conference on Neural Networks (IJCNN)}}}\ (\bibinfo
  {year} {2017})\ pp.\ \bibinfo {pages} {1578--1585}\BibitemShut {NoStop}%
\bibitem [{\citenamefont {Ismail~Fawaz}\ \emph {et~al.}(2019)\citenamefont
  {Ismail~Fawaz}, \citenamefont {Forestier}, \citenamefont {Weber},
  \citenamefont {Idoumghar},\ and\ \citenamefont {Muller}}]{IsmailFawaz2019}%
  \BibitemOpen
  \bibfield  {author} {\bibinfo {author} {\bibfnamefont {H.}~\bibnamefont
  {Ismail~Fawaz}}, \bibinfo {author} {\bibfnamefont {G.}~\bibnamefont
  {Forestier}}, \bibinfo {author} {\bibfnamefont {J.}~\bibnamefont {Weber}},
  \bibinfo {author} {\bibfnamefont {L.}~\bibnamefont {Idoumghar}},\ and\
  \bibinfo {author} {\bibfnamefont {P.-A.}\ \bibnamefont {Muller}},\ }\bibfield
   {title} {\bibinfo {title} {Deep learning for time series classification: A
  review},\ }\href {https://doi.org/10.1007/s10618-019-00619-1} {\bibfield
  {journal} {\bibinfo  {journal} {Data Mining and Knowledge Discovery}\
  }\textbf {\bibinfo {volume} {33}},\ \bibinfo {pages} {917} (\bibinfo {year}
  {2019})}\BibitemShut {NoStop}%
\bibitem [{\citenamefont {Ismail~Fawaz}\ \emph {et~al.}(2020)\citenamefont
  {Ismail~Fawaz}, \citenamefont {Lucas}, \citenamefont {Forestier},
  \citenamefont {Pelletier}, \citenamefont {Schmidt}, \citenamefont {Weber},
  \citenamefont {Webb}, \citenamefont {Idoumghar}, \citenamefont {Muller},\
  and\ \citenamefont {Petitjean}}]{IsmailFawaz2020}%
  \BibitemOpen
  \bibfield  {author} {\bibinfo {author} {\bibfnamefont {H.}~\bibnamefont
  {Ismail~Fawaz}}, \bibinfo {author} {\bibfnamefont {B.}~\bibnamefont {Lucas}},
  \bibinfo {author} {\bibfnamefont {G.}~\bibnamefont {Forestier}}, \bibinfo
  {author} {\bibfnamefont {C.}~\bibnamefont {Pelletier}}, \bibinfo {author}
  {\bibfnamefont {D.~F.}\ \bibnamefont {Schmidt}}, \bibinfo {author}
  {\bibfnamefont {J.}~\bibnamefont {Weber}}, \bibinfo {author} {\bibfnamefont
  {G.~I.}\ \bibnamefont {Webb}}, \bibinfo {author} {\bibfnamefont
  {L.}~\bibnamefont {Idoumghar}}, \bibinfo {author} {\bibfnamefont {P.-A.}\
  \bibnamefont {Muller}},\ and\ \bibinfo {author} {\bibfnamefont
  {F.}~\bibnamefont {Petitjean}},\ }\bibfield  {title} {\bibinfo {title}
  {Inceptiontime: Finding alexnet for time series classification},\ }\href
  {https://doi.org/10.1007/s10618-020-00710-y} {\bibfield  {journal} {\bibinfo
  {journal} {Data Mining and Knowledge Discovery}\ }\textbf {\bibinfo {volume}
  {34}},\ \bibinfo {pages} {1936} (\bibinfo {year} {2020})}\BibitemShut
  {NoStop}%
\bibitem [{\citenamefont {Wen}\ \emph {et~al.}(2023)\citenamefont {Wen},
  \citenamefont {Zhou}, \citenamefont {Zhang}, \citenamefont {Chen},
  \citenamefont {Ma}, \citenamefont {Yan},\ and\ \citenamefont
  {Sun}}]{Wen2023}%
  \BibitemOpen
  \bibfield  {author} {\bibinfo {author} {\bibfnamefont {Q.}~\bibnamefont
  {Wen}}, \bibinfo {author} {\bibfnamefont {T.}~\bibnamefont {Zhou}}, \bibinfo
  {author} {\bibfnamefont {C.}~\bibnamefont {Zhang}}, \bibinfo {author}
  {\bibfnamefont {W.}~\bibnamefont {Chen}}, \bibinfo {author} {\bibfnamefont
  {Z.}~\bibnamefont {Ma}}, \bibinfo {author} {\bibfnamefont {J.}~\bibnamefont
  {Yan}},\ and\ \bibinfo {author} {\bibfnamefont {L.}~\bibnamefont {Sun}},\
  }\bibfield  {title} {\bibinfo {title} {Transformers in time series: A
  survey},\ }in\ \href {https://doi.org/10.24963/ijcai.2023/759} {\emph
  {\bibinfo {booktitle} {Proceedings of the Thirty-Second International Joint
  Conference on Artificial Intelligence}}}\ (\bibinfo {year} {2023})\ pp.\
  \bibinfo {pages} {6778--6786}\BibitemShut {NoStop}%
\bibitem [{\citenamefont {Takens}(2006)}]{takens2006}%
  \BibitemOpen
  \bibfield  {author} {\bibinfo {author} {\bibfnamefont {F.}~\bibnamefont
  {Takens}},\ }\bibfield  {title} {\bibinfo {title} {Detecting strange
  attractors in turbulence},\ }in\ \href
  {https://link.springer.com/chapter/10.1007/BFb0091924} {\emph {\bibinfo
  {booktitle} {Dynamical Systems and Turbulence, Warwick 1980: proceedings of a
  symposium held at the University of Warwick 1979/80}}}\ (\bibinfo
  {organization} {Springer},\ \bibinfo {year} {2006})\ pp.\ \bibinfo {pages}
  {366--381}\BibitemShut {NoStop}%
\bibitem [{\citenamefont {Myers}\ \emph {et~al.}(2019)\citenamefont {Myers},
  \citenamefont {Munch},\ and\ \citenamefont {Khasawneh}}]{Myers2019}%
  \BibitemOpen
  \bibfield  {author} {\bibinfo {author} {\bibfnamefont {A.}~\bibnamefont
  {Myers}}, \bibinfo {author} {\bibfnamefont {E.}~\bibnamefont {Munch}},\ and\
  \bibinfo {author} {\bibfnamefont {F.~A.}\ \bibnamefont {Khasawneh}},\
  }\bibfield  {title} {\bibinfo {title} {Persistent homology of complex
  networks for dynamic state detection},\ }\href
  {https://doi.org/10.1103/PhysRevE.100.022314} {\bibfield  {journal} {\bibinfo
   {journal} {Phys. Rev. E}\ }\textbf {\bibinfo {volume} {100}},\ \bibinfo
  {pages} {022314} (\bibinfo {year} {2019})}\BibitemShut {NoStop}%
\bibitem [{\citenamefont {Tempelman}\ and\ \citenamefont
  {Khasawneh}(2020)}]{Tempelman2020}%
  \BibitemOpen
  \bibfield  {author} {\bibinfo {author} {\bibfnamefont {J.~R.}\ \bibnamefont
  {Tempelman}}\ and\ \bibinfo {author} {\bibfnamefont {F.~A.}\ \bibnamefont
  {Khasawneh}},\ }\bibfield  {title} {\bibinfo {title} {A look into chaos
  detection through topological data analysis},\ }\href
  {https://doi.org/10.1016/j.physd.2020.132446} {\bibfield  {journal} {\bibinfo
   {journal} {Physica D: Nonlinear Phenomena}\ }\textbf {\bibinfo {volume}
  {406}},\ \bibinfo {pages} {132446} (\bibinfo {year} {2020})}\BibitemShut
  {NoStop}%
\bibitem [{\citenamefont {Bubenik}(2015)}]{bubenik2015statistical}%
  \BibitemOpen
  \bibfield  {author} {\bibinfo {author} {\bibfnamefont {P.}~\bibnamefont
  {Bubenik}},\ }\bibfield  {title} {\bibinfo {title} {Statistical topological
  data analysis using persistence landscapes},\ }\href
  {https://www.jmlr.org/papers/v16/bubenik15a.html} {\bibfield  {journal}
  {\bibinfo  {journal} {The Journal of Machine Learning Research}\ }\textbf
  {\bibinfo {volume} {16}},\ \bibinfo {pages} {77} (\bibinfo {year}
  {2015})}\BibitemShut {NoStop}%
\bibitem [{\citenamefont {Adams}\ \emph {et~al.}(2017)\citenamefont {Adams},
  \citenamefont {Emerson}, \citenamefont {Kirby}, \citenamefont {Neville},
  \citenamefont {Peterson}, \citenamefont {Shipman}, \citenamefont
  {Chepushtanova}, \citenamefont {Hanson}, \citenamefont {Motta},\ and\
  \citenamefont {Ziegelmeier}}]{adams2017persistence}%
  \BibitemOpen
  \bibfield  {author} {\bibinfo {author} {\bibfnamefont {H.}~\bibnamefont
  {Adams}}, \bibinfo {author} {\bibfnamefont {T.}~\bibnamefont {Emerson}},
  \bibinfo {author} {\bibfnamefont {M.}~\bibnamefont {Kirby}}, \bibinfo
  {author} {\bibfnamefont {R.}~\bibnamefont {Neville}}, \bibinfo {author}
  {\bibfnamefont {C.}~\bibnamefont {Peterson}}, \bibinfo {author}
  {\bibfnamefont {P.}~\bibnamefont {Shipman}}, \bibinfo {author} {\bibfnamefont
  {S.}~\bibnamefont {Chepushtanova}}, \bibinfo {author} {\bibfnamefont
  {E.}~\bibnamefont {Hanson}}, \bibinfo {author} {\bibfnamefont
  {F.}~\bibnamefont {Motta}},\ and\ \bibinfo {author} {\bibfnamefont
  {L.}~\bibnamefont {Ziegelmeier}},\ }\bibfield  {title} {\bibinfo {title}
  {Persistence images: A stable vector representation of persistent homology},\
  }\href {https://jmlr.org/papers/v18/16-337.html} {\bibfield  {journal}
  {\bibinfo  {journal} {Journal of Machine Learning Research}\ }\textbf
  {\bibinfo {volume} {18}},\ \bibinfo {pages} {1} (\bibinfo {year}
  {2017})}\BibitemShut {NoStop}%
\bibitem [{\citenamefont {Chevyrev}\ \emph {et~al.}(2020)\citenamefont
  {Chevyrev}, \citenamefont {Nanda},\ and\ \citenamefont
  {Oberhauser}}]{chevyrev2018persistence}%
  \BibitemOpen
  \bibfield  {author} {\bibinfo {author} {\bibfnamefont {I.}~\bibnamefont
  {Chevyrev}}, \bibinfo {author} {\bibfnamefont {V.}~\bibnamefont {Nanda}},\
  and\ \bibinfo {author} {\bibfnamefont {H.}~\bibnamefont {Oberhauser}},\
  }\bibfield  {title} {\bibinfo {title} {Persistence paths and signature
  features in topological data analysis},\ }\href
  {https://doi.org/10.1109/TPAMI.2018.2885516} {\bibfield  {journal} {\bibinfo
  {journal} {IEEE Transactions on Pattern Analysis and Machine Intelligence}\
  }\textbf {\bibinfo {volume} {42}},\ \bibinfo {pages} {192} (\bibinfo {year}
  {2020})}\BibitemShut {NoStop}%
\bibitem [{\citenamefont {Atienza}\ \emph {et~al.}(2020)\citenamefont
  {Atienza}, \citenamefont {Gonzalez-Diaz},\ and\ \citenamefont
  {Soriano-Trigueros}}]{atienza2020stability}%
  \BibitemOpen
  \bibfield  {author} {\bibinfo {author} {\bibfnamefont {N.}~\bibnamefont
  {Atienza}}, \bibinfo {author} {\bibfnamefont {R.}~\bibnamefont
  {Gonzalez-Diaz}},\ and\ \bibinfo {author} {\bibfnamefont {M.}~\bibnamefont
  {Soriano-Trigueros}},\ }\bibfield  {title} {\bibinfo {title} {On the
  stability of persistent entropy and new summary functions for topological
  data analysis},\ }\href {https://doi.org/10.1016/j.patcog.2020.107509}
  {\bibfield  {journal} {\bibinfo  {journal} {Pattern Recognition}\ }\textbf
  {\bibinfo {volume} {107}},\ \bibinfo {pages} {107509} (\bibinfo {year}
  {2020})}\BibitemShut {NoStop}%
\bibitem [{\citenamefont {Umeda}(2017)}]{umeda2017time}%
  \BibitemOpen
  \bibfield  {author} {\bibinfo {author} {\bibfnamefont {Y.}~\bibnamefont
  {Umeda}},\ }\bibfield  {title} {\bibinfo {title} {Time series classification
  via topological data analysis},\ }\href {https://doi.org/10.1527/tjsai.D-G72}
  {\bibfield  {journal} {\bibinfo  {journal} {Transactions of the Japanese
  Society for Artificial Intelligence}\ }\textbf {\bibinfo {volume} {32}},\
  \bibinfo {pages} {D} (\bibinfo {year} {2017})}\BibitemShut {NoStop}%
\bibitem [{\citenamefont {Karan}\ and\ \citenamefont
  {Kaygun}(2021)}]{Karan2021}%
  \BibitemOpen
  \bibfield  {author} {\bibinfo {author} {\bibfnamefont {A.}~\bibnamefont
  {Karan}}\ and\ \bibinfo {author} {\bibfnamefont {A.}~\bibnamefont {Kaygun}},\
  }\bibfield  {title} {\bibinfo {title} {Time series classification via
  topological data analysis},\ }\href
  {https://doi.org/10.1016/j.eswa.2021.115326} {\bibfield  {journal} {\bibinfo
  {journal} {Expert Systems with Applications}\ }\textbf {\bibinfo {volume}
  {183}},\ \bibinfo {pages} {115326} (\bibinfo {year} {2021})}\BibitemShut
  {NoStop}%
\bibitem [{\citenamefont {Khasawneh}\ \emph {et~al.}(2025)\citenamefont
  {Khasawneh}, \citenamefont {Munch}, \citenamefont {Barnes}, \citenamefont
  {Chumley}, \citenamefont {G{\"u}zel}, \citenamefont {Myers}, \citenamefont
  {Tanweer}, \citenamefont {Tymochko},\ and\ \citenamefont
  {Yesilli}}]{Khasawneh2025}%
  \BibitemOpen
  \bibfield  {author} {\bibinfo {author} {\bibfnamefont {F.~A.}\ \bibnamefont
  {Khasawneh}}, \bibinfo {author} {\bibfnamefont {E.}~\bibnamefont {Munch}},
  \bibinfo {author} {\bibfnamefont {D.}~\bibnamefont {Barnes}}, \bibinfo
  {author} {\bibfnamefont {M.~M.}\ \bibnamefont {Chumley}}, \bibinfo {author}
  {\bibfnamefont {{\.I}.}~\bibnamefont {G{\"u}zel}}, \bibinfo {author}
  {\bibfnamefont {A.~D.}\ \bibnamefont {Myers}}, \bibinfo {author}
  {\bibfnamefont {S.}~\bibnamefont {Tanweer}}, \bibinfo {author} {\bibfnamefont
  {S.}~\bibnamefont {Tymochko}},\ and\ \bibinfo {author} {\bibfnamefont
  {M.}~\bibnamefont {Yesilli}},\ }\bibfield  {title} {\bibinfo {title}
  {Teaspoon: A python package for topological signal processing},\ }\href
  {https://doi.org/10.21105/joss.07243} {\bibfield  {journal} {\bibinfo
  {journal} {Journal of Open Source Software}\ }\textbf {\bibinfo {volume}
  {10}},\ \bibinfo {pages} {7243} (\bibinfo {year} {2025})}\BibitemShut
  {NoStop}%
\bibitem [{\citenamefont {Lorenz}(1963)}]{lorenz2017}%
  \BibitemOpen
  \bibfield  {author} {\bibinfo {author} {\bibfnamefont {E.~N.}\ \bibnamefont
  {Lorenz}},\ }\bibfield  {title} {\bibinfo {title} {Deterministic nonperiodic
  flow},\ }\href {https://doi.org/10.1175/1520-0469(1963)020<0130:DNF>2.0.CO;2}
  {\bibfield  {journal} {\bibinfo  {journal} {Journal of Atmospheric Sciences}\
  }\textbf {\bibinfo {volume} {20}},\ \bibinfo {pages} {130 } (\bibinfo {year}
  {1963})}\BibitemShut {NoStop}%
\bibitem [{\citenamefont {H{\'e}non}\ and\ \citenamefont
  {Heiles}(1964)}]{henon1964}%
  \BibitemOpen
  \bibfield  {author} {\bibinfo {author} {\bibfnamefont {M.}~\bibnamefont
  {H{\'e}non}}\ and\ \bibinfo {author} {\bibfnamefont {C.}~\bibnamefont
  {Heiles}},\ }\bibfield  {title} {\bibinfo {title} {The applicability of the
  third integral of motion: some numerical experiments},\ }\href
  {https://doi.org/https://doi.org/10.1086%2F109234} {\bibfield  {journal}
  {\bibinfo  {journal} {Astronomical Journal, Vol. 69, p. 73 (1964)}\ }\textbf
  {\bibinfo {volume} {69}},\ \bibinfo {pages} {73} (\bibinfo {year}
  {1964})}\BibitemShut {NoStop}%
\bibitem [{\citenamefont {Grassberger}\ and\ \citenamefont
  {Procaccia}(1983{\natexlab{b}})}]{grassberger1983}%
  \BibitemOpen
  \bibfield  {author} {\bibinfo {author} {\bibfnamefont {P.}~\bibnamefont
  {Grassberger}}\ and\ \bibinfo {author} {\bibfnamefont {I.}~\bibnamefont
  {Procaccia}},\ }\bibfield  {title} {\bibinfo {title} {Measuring the
  strangeness of strange attractors},\ }\href
  {https://doi.org/https://doi.org/10.1016/0167-2789(83)90298-1} {\bibfield
  {journal} {\bibinfo  {journal} {Physica D: nonlinear phenomena}\ }\textbf
  {\bibinfo {volume} {9}},\ \bibinfo {pages} {189} (\bibinfo {year}
  {1983}{\natexlab{b}})}\BibitemShut {NoStop}%
\bibitem [{\citenamefont {Theiler}(1987)}]{Theiler1987}%
  \BibitemOpen
  \bibfield  {author} {\bibinfo {author} {\bibfnamefont {J.}~\bibnamefont
  {Theiler}},\ }\bibfield  {title} {\bibinfo {title} {Efficient algorithm for
  estimating the correlation dimension from a set of discrete points},\ }\href
  {https://doi.org/10.1103/PhysRevA.36.4456} {\bibfield  {journal} {\bibinfo
  {journal} {Phys. Rev. A}\ }\textbf {\bibinfo {volume} {36}},\ \bibinfo
  {pages} {4456} (\bibinfo {year} {1987})}\BibitemShut {NoStop}%
\bibitem [{Note1()}]{Note1}%
  \BibitemOpen
  \bibinfo {note} {For computational reasons, we set $H_{\protect \rm max}=2$
  and therefore retain $H_0$ and $H_1$.}\BibitemShut {Stop}%
\bibitem [{Note2()}]{Note2}%
  \BibitemOpen
  \bibinfo {note} {Because the capped death value depends on the largest finite
  death observed for a given point cloud, the resulting lifetime assigned to
  the essential $H_0$ class is an implementation-dependent finite value and
  should not be interpreted as a topological invariant of the underlying
  attractor.}\BibitemShut {Stop}%
\bibitem [{\citenamefont {Atienza}\ \emph {et~al.}(2019)\citenamefont
  {Atienza}, \citenamefont {Gonzalez-Diaz},\ and\ \citenamefont
  {Rucco}}]{atienza2019persistent}%
  \BibitemOpen
  \bibfield  {author} {\bibinfo {author} {\bibfnamefont {N.}~\bibnamefont
  {Atienza}}, \bibinfo {author} {\bibfnamefont {R.}~\bibnamefont
  {Gonzalez-Diaz}},\ and\ \bibinfo {author} {\bibfnamefont {M.}~\bibnamefont
  {Rucco}},\ }\bibfield  {title} {\bibinfo {title} {Persistent entropy for
  separating topological features from noise in vietoris-rips complexes},\
  }\href {https://doi.org/https://doi.org/10.1007/s10844-017-0473-4} {\bibfield
   {journal} {\bibinfo  {journal} {Journal of Intelligent Information Systems}\
  }\textbf {\bibinfo {volume} {52}},\ \bibinfo {pages} {637} (\bibinfo {year}
  {2019})}\BibitemShut {NoStop}%
\bibitem [{Note3()}]{Note3}%
  \BibitemOpen
  \bibinfo {note} {The factor $0.05$ defines a fixed relative threshold used
  throughout the analysis. Here, ``significant'' refers only to this
  thresholding criterion and does not imply statistical
  significance.}\BibitemShut {Stop}%
\bibitem [{\citenamefont {Katharopoulos}\ \emph {et~al.}(2020)\citenamefont
  {Katharopoulos}, \citenamefont {Vyas}, \citenamefont {Pappas},\ and\
  \citenamefont {Fleuret}}]{Katharopoulos2020}%
  \BibitemOpen
  \bibfield  {author} {\bibinfo {author} {\bibfnamefont {A.}~\bibnamefont
  {Katharopoulos}}, \bibinfo {author} {\bibfnamefont {A.}~\bibnamefont {Vyas}},
  \bibinfo {author} {\bibfnamefont {N.}~\bibnamefont {Pappas}},\ and\ \bibinfo
  {author} {\bibfnamefont {F.}~\bibnamefont {Fleuret}},\ }\bibfield  {title}
  {\bibinfo {title} {Transformers are rnns: Fast autoregressive transformers
  with linear attention},\ }in\ \href
  {https://proceedings.mlr.press/v119/katharopoulos20a/katharopoulos20a.pdf}
  {\emph {\bibinfo {booktitle} {Proceedings of the 37th International
  Conference on Machine Learning}}},\ Vol.\ \bibinfo {volume} {119}\ (\bibinfo
  {publisher} {PMLR},\ \bibinfo {year} {2020})\ pp.\ \bibinfo {pages}
  {5156--5165}\BibitemShut {NoStop}%
\bibitem [{Note4()}]{Note4}%
  \BibitemOpen
  \bibinfo {note} {The exponential linear unit reads $\displaystyle {\protect
  \rm ELU}(x)=x $ for $x>0$ and $\displaystyle {\protect \rm ELU}(x)=e^x-1 $
  for $x\leq 0$.}\BibitemShut {Stop}%
\bibitem [{\citenamefont {Kingma}\ and\ \citenamefont
  {Ba}(2014)}]{kingma2014adam}%
  \BibitemOpen
  \bibfield  {author} {\bibinfo {author} {\bibfnamefont {D.~P.}\ \bibnamefont
  {Kingma}}\ and\ \bibinfo {author} {\bibfnamefont {J.}~\bibnamefont {Ba}},\
  }\bibfield  {title} {\bibinfo {title} {Adam: A method for stochastic
  optimization},\ }\bibfield  {journal} {\bibinfo  {journal} {arXiv preprint
  arXiv:1412.6980}\ }\href {https://doi.org/10.48550/arXiv.1412.6980}
  {10.48550/arXiv.1412.6980} (\bibinfo {year} {2014})\BibitemShut {NoStop}%
\bibitem [{\citenamefont {Loshchilov}\ and\ \citenamefont
  {Hutter}(2017)}]{loshchilov2017adamw}%
  \BibitemOpen
  \bibfield  {author} {\bibinfo {author} {\bibfnamefont {I.}~\bibnamefont
  {Loshchilov}}\ and\ \bibinfo {author} {\bibfnamefont {F.}~\bibnamefont
  {Hutter}},\ }\bibfield  {title} {\bibinfo {title} {Decoupled weight decay
  regularization},\ }\href@noop {} {\bibfield  {journal} {\bibinfo  {journal}
  {arXiv preprint arXiv:1711.05101}\ } (\bibinfo {year} {2017})}\BibitemShut
  {NoStop}%
\bibitem [{\citenamefont {Guo}\ \emph {et~al.}(2017)\citenamefont {Guo},
  \citenamefont {Pleiss}, \citenamefont {Sun},\ and\ \citenamefont
  {Weinberger}}]{guo2017calibration}%
  \BibitemOpen
  \bibfield  {author} {\bibinfo {author} {\bibfnamefont {C.}~\bibnamefont
  {Guo}}, \bibinfo {author} {\bibfnamefont {G.}~\bibnamefont {Pleiss}},
  \bibinfo {author} {\bibfnamefont {Y.}~\bibnamefont {Sun}},\ and\ \bibinfo
  {author} {\bibfnamefont {K.~Q.}\ \bibnamefont {Weinberger}},\ }\bibfield
  {title} {\bibinfo {title} {On calibration of modern neural networks},\ }in\
  \href@noop {} {\emph {\bibinfo {booktitle} {Proceedings of the 34th
  International Conference on Machine Learning - Volume 70}}},\ \bibinfo
  {series and number} {ICML'17}\ (\bibinfo  {publisher} {JMLR.org},\ \bibinfo
  {year} {2017})\ p.\ \bibinfo {pages} {1321–1330}\BibitemShut {NoStop}%
\bibitem [{\citenamefont {Eckmann}\ \emph {et~al.}(1986)\citenamefont
  {Eckmann}, \citenamefont {Kamphorst}, \citenamefont {Ruelle},\ and\
  \citenamefont {Ciliberto}}]{Eckmann1986}%
  \BibitemOpen
  \bibfield  {author} {\bibinfo {author} {\bibfnamefont {J.~P.}\ \bibnamefont
  {Eckmann}}, \bibinfo {author} {\bibfnamefont {S.~O.}\ \bibnamefont
  {Kamphorst}}, \bibinfo {author} {\bibfnamefont {D.}~\bibnamefont {Ruelle}},\
  and\ \bibinfo {author} {\bibfnamefont {S.}~\bibnamefont {Ciliberto}},\
  }\bibfield  {title} {\bibinfo {title} {Liapunov exponents from time series},\
  }\href {https://doi.org/10.1103/PhysRevA.34.4971} {\bibfield  {journal}
  {\bibinfo  {journal} {Phys. Rev. A}\ }\textbf {\bibinfo {volume} {34}},\
  \bibinfo {pages} {4971} (\bibinfo {year} {1986})}\BibitemShut {NoStop}%
\bibitem [{Note5()}]{Note5}%
  \BibitemOpen
  \bibinfo {note} {The single essential $H_{0}$ class, whose death time is
  formally infinite, is capped at the largest finite death value observed
  across all tracked homology dimensions for that point cloud before any
  downstream statistics are computed; this ensures every reported quantity,
  including the entropy, lifetime, Betti-curve, and persistence-image features,
  is finite-valued by construction.}\BibitemShut {Stop}%
\bibitem [{Note6()}]{Note6}%
  \BibitemOpen
  \bibinfo {note} {Because Betti curves are piecewise constant, this criterion
  does not count turning points separated by one or more zero-valued
  differences (plateaus); it therefore provides a conservative lower bound on
  the number of true direction changes in the curve, rather than an exhaustive
  count.}\BibitemShut {Stop}%
\bibitem [{\citenamefont {Jordan}\ \emph {et~al.}(2024)\citenamefont {Jordan},
  \citenamefont {Jin}, \citenamefont {Boza}, \citenamefont {Jiacheng},
  \citenamefont {Cesista}, \citenamefont {Newhouse},\ and\ \citenamefont
  {Bernstein}}]{jordan2024muon}%
  \BibitemOpen
  \bibfield  {author} {\bibinfo {author} {\bibfnamefont {K.}~\bibnamefont
  {Jordan}}, \bibinfo {author} {\bibfnamefont {Y.}~\bibnamefont {Jin}},
  \bibinfo {author} {\bibfnamefont {V.}~\bibnamefont {Boza}}, \bibinfo {author}
  {\bibfnamefont {Y.}~\bibnamefont {Jiacheng}}, \bibinfo {author}
  {\bibfnamefont {F.}~\bibnamefont {Cesista}}, \bibinfo {author} {\bibfnamefont
  {L.}~\bibnamefont {Newhouse}},\ and\ \bibinfo {author} {\bibfnamefont
  {J.}~\bibnamefont {Bernstein}},\ }\href
  {https://kellerjordan.github.io/posts/muon/} {\bibinfo {title} {Muon: An
  optimizer for hidden layers in neural networks}} (\bibinfo {year}
  {2024})\BibitemShut {NoStop}%
\bibitem [{\citenamefont {Chazal}\ \emph {et~al.}(2014)\citenamefont {Chazal},
  \citenamefont {de~Silva},\ and\ \citenamefont
  {Oudot}}]{chazal2014persistence}%
  \BibitemOpen
  \bibfield  {author} {\bibinfo {author} {\bibfnamefont {F.}~\bibnamefont
  {Chazal}}, \bibinfo {author} {\bibfnamefont {V.}~\bibnamefont {de~Silva}},\
  and\ \bibinfo {author} {\bibfnamefont {S.}~\bibnamefont {Oudot}},\ }\bibfield
   {title} {\bibinfo {title} {Persistence stability for geometric complexes},\
  }\href {https://doi.org/10.1007/s10711-013-9937-z} {\bibfield  {journal}
  {\bibinfo  {journal} {Geometriae Dedicata}\ }\textbf {\bibinfo {volume}
  {173}},\ \bibinfo {pages} {193} (\bibinfo {year} {2014})}\BibitemShut
  {NoStop}%
\bibitem [{\citenamefont {Cao}\ \emph {et~al.}(2023)\citenamefont {Cao},
  \citenamefont {Leykam},\ and\ \citenamefont {Angelakis}}]{Cao2023}%
  \BibitemOpen
  \bibfield  {author} {\bibinfo {author} {\bibfnamefont {H.}~\bibnamefont
  {Cao}}, \bibinfo {author} {\bibfnamefont {D.}~\bibnamefont {Leykam}},\ and\
  \bibinfo {author} {\bibfnamefont {D.~G.}\ \bibnamefont {Angelakis}},\
  }\bibfield  {title} {\bibinfo {title} {Unravelling quantum chaos using
  persistent homology},\ }\href {https://doi.org/10.1103/PhysRevE.107.044204}
  {\bibfield  {journal} {\bibinfo  {journal} {Phys. Rev. E}\ }\textbf {\bibinfo
  {volume} {107}},\ \bibinfo {pages} {044204} (\bibinfo {year}
  {2023})}\BibitemShut {NoStop}%
\bibitem [{\citenamefont {Kerem~Maden}\ \emph {et~al.}(2026)\citenamefont
  {Kerem~Maden}, \citenamefont {Ullah}, \citenamefont {Coskunuzer},\ and\
  \citenamefont {Mustecaplıoglu}}]{KeremMaden_2026}%
  \BibitemOpen
  \bibfield  {author} {\bibinfo {author} {\bibfnamefont {M.}~\bibnamefont
  {Kerem~Maden}}, \bibinfo {author} {\bibfnamefont {A.}~\bibnamefont {Ullah}},
  \bibinfo {author} {\bibfnamefont {B.}~\bibnamefont {Coskunuzer}},\ and\
  \bibinfo {author} {\bibfnamefont {Q.~E.}\ \bibnamefont {Mustecaplıoglu}},\
  }\bibfield  {title} {\bibinfo {title} {Topological engine monitor: persistent
  homology-based fault detection in finite-time quantum engines},\ }\href
  {https://doi.org/10.1088/2058-9565/ae98eb} {\bibfield  {journal} {\bibinfo
  {journal} {Quantum Science and Technology}\ }\textbf {\bibinfo {volume}
  {11}},\ \bibinfo {pages} {045012} (\bibinfo {year} {2026})}\BibitemShut
  {NoStop}%
\bibitem [{\citenamefont {Tralie}\ \emph {et~al.}(2018)\citenamefont {Tralie},
  \citenamefont {Saul},\ and\ \citenamefont {Bar-On}}]{ctralie2018ripser}%
  \BibitemOpen
  \bibfield  {author} {\bibinfo {author} {\bibfnamefont {C.}~\bibnamefont
  {Tralie}}, \bibinfo {author} {\bibfnamefont {N.}~\bibnamefont {Saul}},\ and\
  \bibinfo {author} {\bibfnamefont {R.}~\bibnamefont {Bar-On}},\ }\bibfield
  {title} {\bibinfo {title} {{Ripser.py}: A lean persistent homology library
  for python},\ }\href {https://doi.org/10.21105/joss.00925} {\bibfield
  {journal} {\bibinfo  {journal} {The Journal of Open Source Software}\
  }\textbf {\bibinfo {volume} {3}},\ \bibinfo {pages} {925} (\bibinfo {year}
  {2018})}\BibitemShut {NoStop}%
\bibitem [{\citenamefont {Saul}\ and\ \citenamefont
  {Tralie}(2019)}]{scikittda2019}%
  \BibitemOpen
  \bibfield  {author} {\bibinfo {author} {\bibfnamefont {N.}~\bibnamefont
  {Saul}}\ and\ \bibinfo {author} {\bibfnamefont {C.}~\bibnamefont {Tralie}},\
  }\href {https://doi.org/10.5281/zenodo.2533369} {\bibinfo {title}
  {Scikit-tda: Topological data analysis for python}} (\bibinfo {year}
  {2019})\BibitemShut {NoStop}%
\bibitem [{\citenamefont {Dlotko}(2022)}]{gudhi:PersistenceRepresentations}%
  \BibitemOpen
  \bibfield  {author} {\bibinfo {author} {\bibfnamefont {P.}~\bibnamefont
  {Dlotko}},\ }\bibfield  {title} {\bibinfo {title} {Persistence
  representations},\ }in\ \href
  {https://gudhi.inria.fr/doc/3.5.0/group___persistence__representations.html}
  {\emph {\bibinfo {booktitle} {{GUDHI} User and Reference Manual}}}\ (\bibinfo
   {publisher} {{GUDHI Editorial Board}},\ \bibinfo {year} {2022})\ \bibinfo
  {edition} {{3.5.0}}\ ed.\BibitemShut {Stop}%
\end{thebibliography}%

 \end{document}